\documentclass[
    sigconf = true,     
    review = false,      
    screen = true,      
    anonymous = false,   
]
{acmart}
\renewcommand\footnotetextcopyrightpermission[1]{} 
\usepackage{amsmath}
\usepackage{mathtools}
\usepackage{xspace}
\usepackage{booktabs}
\usepackage{xcolor}
\usepackage{colortbl}
\usepackage{multirow}
\usepackage{array}
\usepackage{hyperref}
\usepackage{url}
\usepackage{makecell}
\usepackage{changepage}

\usepackage{float}
\usepackage{booktabs}
\usepackage{longtable}
\usepackage{nicefrac}

\usepackage{xspace}
\usepackage{mathtools}
\usepackage{diagbox}
\usepackage{graphics}
\usepackage{rotating}
\usepackage[algo2e,ruled,vlined,linesnumbered]{algorithm2e}
\newcommand{\assign}{\leftarrow}
\usepackage{graphicx}
\usepackage{subfigure}
\usepackage{ifthen}
\usepackage{wrapfig}
\usepackage{balance} 
\usepackage{epsfig}
\usepackage{xcolor, colortbl}
\usepackage[normalem]{ulem}
\usepackage{dsfont}

\newcommand{\oea}{$(1 + 1)$~EA$_{>0}$\xspace}

\newcommand{\om}{\textsc{OneMax}\xspace}
\newcommand{\onemax}{\om}
\newcommand{\lo}{\textsc{LeadingOnes}\xspace}
\newcommand{\jump}{\textsc{Jump}\xspace}

\newcommand{\leadingones}{\lo}

\usepackage{graphicx}
\usepackage[utf8]{inputenc}

\copyrightyear{2021}
\acmYear{2021}
\setcopyright{none}
\acmConference[GECCO '21]{Genetic and Evolutionary Computation Conference}{July 10--14, 2020}{Lille, France}

\begin{document}

\title{Leveraging Benchmarking Data for Informed One-Shot Dynamic Algorithm Selection}

\author{Furong Ye}
\affiliation{%
   \institution{LIACS, Leiden University}
   \city{Leiden}
   \country{Netherlands}}

\author{Carola Doerr}
\affiliation{%
   \institution{Sorbonne Universit\'e, CNRS, LIP6}
   \city{Paris}
   \country{France}}

\author{Thomas B\"ack}
\affiliation{%
   \institution{LIACS, Leiden University}
   \city{Leiden}
   \country{Netherlands}}

\begin{abstract} 
A key challenge in the application of evolutionary algorithms in practice is the selection of an algorithm instance that best suits the problem at hand. What complicates this decision further is that different algorithms may be best suited for different stages of the optimization process. Dynamic algorithm selection and configuration are  therefore well-researched topics in evolutionary computation. However, while hyper-heuristics and parameter control studies typically assume a setting in which the algorithm needs to be chosen while running the algorithms, without prior information, AutoML approaches such as hyper-parameter tuning and automated algorithm configuration assume the possibility of evaluating different configurations before making a final recommendation. 
In practice, however, we are often in a middle-ground between these two settings, where we need to decide on the algorithm instance before the run (``oneshot'' setting), but where we have (possibly lots of) data available on which we can base an informed decision.

We analyze in this work how such prior performance data can be used to infer informed dynamic algorithm selection schemes for the solution of pseudo-Boolean optimization problems. Our specific use-case considers a family of genetic algorithms. 
       
\end{abstract}



\maketitle
\fancyhf{}
\renewcommand\headrulewidth{0pt}
\fancyhead[REO]{Furong Ye, Carola Doerr, and Thomas B\"ack}
\section{Introduction}
It is well-known that genetic algorithms (GAs) require proper parameter settings and operators to work efficiently. Though many parameter control methods \cite{bartz2005sequential,birattari2010f,SMAC,ParamILS,lopez2016irace,van2019automatic} have been proposed to tune the parameters of algorithms, they usually provide only a fixed suggestion for the algorithms. However, recent research has shown that the optimal settings of parameters may adjust at different optimization stages, so searching for an optimal static parameter setting will prevent us from identifying the best solver.  Simultaneously, self-adaptation has been studied for tuning parameters on the fly, but the effectiveness of an adaptive method also depends on the properties of the problem and the stage of optimization. For example, the optimal mutation rate of $(1+\lambda)$~EA has been proven to be non-static for \onemax, and a $(1+\lambda)$~EA$_{r/2,2r}$ with self-adaptive mutation rate has been proposed in \cite{doerr20191+}.  $(1+\lambda)$~EA$_{r/2,2r}$ has shown its ability of following the dynamic optimal mutation rate for \onemax in~\cite{doerr20191+}. However, when we solve the \jump function, $(1+\lambda)$~EA$_{r/2,2r}$ will not be the best choice anymore. 

\jump is similar to \onemax, but the values in the interval $[n-m+1,n-1]$ are either set to zero or to $n-\onemax(x)$ (both variants are studied in the literature, see~\cite{JansenREALjump} for a discussion and yet another jump function) so that in order to reach the global optimum, elitist algorithms like the $(1+\lambda)$~EA need to jump from a solution of fitness $n-m$ directly to the optimum. While the $(1+\lambda)$~EA$_{r/2,2r}$ is still efficient at the stage before reaching fitness layer $n-m$, it is not very good at jumping to the optimum. For this last step, other methods, including crossover-based algorithms  \cite{whitley2018exploration,dang2017escaping} may be a better choice.

To tackle situations as above, we would, intuitively, want to select a best-suited algorithm for each stage of the optimization process. This idea defines a new meta-optimization problem, which is called the \textit{dynamic algorithm selection} (dynAS) problem. The dynAS is expected to unlock the potential benefit from switching among different algorithms online. Related work has been performed on black-box optimization for numeric optimization \cite{vermetten2020towards}. Based on the rich BBOB data set~\cite{BBOBdata}, \cite{vermetten2020towards} investigates the potential improvement that can be achieved from switching between using solvers. However, the results presented in~\cite{vermetten2020towards} are restricted to a theoretical assessment, without an experimental proof.

The dynAS approach is closely related to hyper-heuristics~\cite{BurkeGHKOOQ13hyper,pillay2018hyper} and Algorithm control~\cite{biedenkapp2020dynamic}. 
However, while hyper-heuristics and parameter control studies typically assume a setting in which the algorithm needs to be chosen while running the algorithms, without prior information (\emph{``on-the-fly''}, \emph{``online''}, or \emph{``adaptive''} selection), 
AutoML approaches such as hyper-parameter tuning and automated algorithm configuration assume the possibility of evaluating different configurations before making a final recommendation (\emph{``offline''} tuning). 
In practice, however, we are often in a middle-ground between these two settings, where we need to decide on the algorithm instance before the run (``one-shot'' decision, no training or partial evaluations possible), but where we have (possibly lots of) data available on which we can base our decision (the \emph{``informed''} setting).

\textbf{Our contribution:} 
We analyze in this work how well existing benchmark data can be used for the selection of suitable algorithm combinations. We base our experiments on the results of the benchmark study presented in~\cite{Yeppsn2020}. This dataset provides us with detailed performance records for 80 different instances of a family of $(\mu+\lambda)$ GAs run on 25 pseudo-Boolean problems introduced in \cite{doerr2020benchmarking} (first 23 functions) and \cite{Yeppsn2020} (last two problems). The data records are stored in a COCO-like format~\cite{hansen2020coco} and are conveniently interpretable by IOHanalyzer~\cite{IOHanalyzer}, the data analysis and visualization module of IOHprofiler~\cite{doerr2018iohprofiler}. 

Starting with an assessment of the performance improvement that we can expect from using the dynAS, extensive experimentation has been performed to reveal the effectiveness of the dynAS and difficulties we may encounter for future study. dynAS is a hard problem, and we cannot fully solve it in this work. So instead of proposing a complete solution for it, our work highlights the advantages of dynAS in some settings, and experimental results have shown that we can obtain better solvers and spot competitive algorithms for different stages of the optimization process by using the dynAS.

After a purely theoretical investigation of the benchmark data of~\cite{Yeppsn2020}, we expected to gain improvements on all problems. In practice, however, these predicted potentials could not be realized on all problems. We analyze both successful and unsuccessful trials of the dynAS by considering \textit{the set of algorithms}, the \textit{switching points}, and the \textit{properties of the problems (local optima)}, which helps designing solutions of the dynAS in future work.

Simultaneously, we highlight competitive algorithms for different stages of the optimization process on problems, such as \leadingones and W-model~\cite{weise2018difficult} problems, which illustrates why we recommend our work of the dynAS together with benchmarking. 
The reason is that we not only obtain better solvers but also study how algorithms perform at stages on different problems. By applying the dynAS for benchmarking, we can easily spot useful combinations of switching algorithms from the set of possible combinations,
which also builds a bridge from practical experiment to theoretical analysis.
Overall, we promote the idea of dynamic genetic algorithm selection in this work. By applying the dynAS, we obtain better solvers for most of the IOHprofiler benchmark problems, and we address the main challenges of the dynAS for future study. Moreover, we highlight the competitive settings 
of the GA for problems such as \leadingones and W-model problems.

\textbf{Outline of the paper:} We recall a formalization of the dynAS problem in Sec.~\ref{sec:prelim}, summarize the GA family and the benchmark problems in Sec.~\ref{sec:3}, demonstrate the effectiveness of dynAS in Sec.~\ref{sec:LO}, discuss its generalization to the IOHprofiler problems in Sec.~\ref{sec:IOH}, and we conclude the paper in Sec.~\ref{sec:conclusions}.

\section{Preliminaries}\label{sec:prelim}
\subsection{Dynamic Algorithm Selection}
The \textit{algorithm selection} problem is to find the best algorithm $A^*$ from an algorithm set $\mathcal{A}$ to solve a problem $P$ \cite{rice1976algorithm}. We call this classic version the \textit{static} algorithm selection, and the definition is given below.

\begin{definition}[AS: Static Algorithm Selection] Given a problem $P$, a set $\mathcal{A}=\{A_1,...,A_n\}$ of algorithms, and a cost metric $c$: $A \times P \mapsto \mathcal{R}$ (e.g., the expected running time to solve the problem), the objective is to find:
$$
A^* \in \underset{A \in \mathcal{A}}{\arg \min }\ c(A,P) 
$$
\end{definition}

For the dynamic algorithm selection (dynAS) discussed in this work, a dynamic selection policy $\pi \in \Pi$ is introduced to define the dynamic method of  algorithm switching. We define the dynAS as below, which refers to the definition of the dynamic algorithm configuration task (dynAC) in \cite{biedenkapp2020dynamic}.

\begin{definition}[dynAS: Dynamic Algorithm Selection] Given a problem $P$, a set $\mathcal{A}=\{A_1,...,A_n\}$ of algorithms, a state description $s_t \in \mathcal{S}$ of solving $P$ at time point $t$, and a cost metric $c$: $\Pi \times P \mapsto \mathcal{R}$ accessing the cost of a dynamic selection policy $\pi$ on a problem $p$ (e.g., the expected running time to solve the problem), the objective is to find a policy $\pi ^ *$: $\mathcal{S} \times P \mapsto \mathcal{A}$, that selects an algorithm $A \in \mathcal{A}$ at time point $t$, by optimizing its cost on the problem $P$:
$$
\pi^* \in \underset{\pi \in \Pi}{\arg \min }\ c(\pi,P).
$$
\end{definition}

We note that Definition 2.1.2 may suggest that the optimal policy may depend on the time elapsed. In practice, however, other indicators such as solution quality are also considered~\cite{vermetten2020towards,VermettenCMAdynAS} or even known to be optimal~\cite{DoerrDY20,BottcherDN10}.  

\subsection{Performance Measure}
\label{sec:ERT-dynAS}
To solve the dynAS problem, we need to define the state description $s_t \in \mathcal{S}$ at time point $t$ and the cost metric $c$. The fixed-target approach of measuring algorithm performance by using the expected running time (ERT) perfectly matches this requirement. The ERT of an algorithm $A$ hitting a target $\phi$ on a problem $P$ is given as below~\cite{hansen2020coco}:
\begin{equation}
\text{ERT}(A,P,\phi) = \frac{\sum_{i=1}^{r}\min\{t_i(A,P,\phi),B\}}{\sum_{i=1}^{r}\mathds{1}\{t_i(A,P,\phi)< \infty\}} \; ,
\label{formula:ERT}
\end{equation}
where $r$ is the number of runs of the algorithm $A$, $B$ is the maximal budget (e.g., maximal number of function evaluation) of the algorithm $A$ on the problem $P$.
\\

\noindent \textbf{Application of ERT} By using the ERT as the cost metric $c$ of the AS problem, we can define the best static algorithm $A^*$ as below:

\begin{definition}[BSA: Best Static Algorithm] Given a problem $P$ and a set $\mathcal{A} = \{A_1, ...,A_n\}$ of algorithms, the best static algorithm $A^*$ for the target $\phi$ is
$$
A^* = \underset{A \in \mathcal{A}}{\arg \min } \; \text{ERT} (A,P,\phi).
$$
\end{definition}

As for the dynAS, we restrict our attention in this work to dynamic selection policies $\pi$ which switch only once, i.e., from using an algorithm $A_1$ to an algorithm $A_2$ at the state $s$, where a fitness $f\ge \phi_s$ is found for the first time. Therefore, the policy $\pi$ can be constructed as $\pi = (A_1,A_2, \phi_s)$ for this \textit{switch-once} dynAS, where $A_1, A_2 \in \mathcal{A}$ and $\phi_s$ is within the domain of fitness values. Then the predicted performance of $\pi$ hitting the final target $\phi_f$ on a problem $P$ can be calculated as:
\begin{equation}
T(\pi, P, \phi_f) = \text{ERT}(A_1,P,\phi_s) + \text{ERT}(A_2,P,\phi_f) - \text{ERT}(A_2,P,\phi_s).
\label{formula:pre-ERT}
\end{equation}

By using the upper predicted performance as the cost metric $c$ of the dynAS problem, we define the best dynamic algorithm selection policy $\pi^*$ as below:
\begin{definition}[BDA: Best Dynamic Algorithm Selection Policy] Given a problem P, a set $\Phi$ of targets, and a set $\mathcal{A} = \{A_1, ...,A_n\}$ of algorithms, the best dynamic algorithm selection policy $\pi^*$ for the target $\phi_f$ is
$$
(A_1, A_2, \phi_s)^* = \pi^*
=   \underset{ \pi \in (\mathcal{A} \times \mathcal{A} \times \Phi)}{\arg \min } T(\pi, P, \phi_f),
$$
where $A_1, A_2 \in \mathcal{A}, \phi_s \in \Phi$.
\end{definition}

\section{Algorithm and Benchmark}\label{sec:3}
We describe in this section a configurable GA framework, the IOHprofiler problems, and our prior benchmark data for the dynAS problem in Sec.~\ref{dynAS-PBO}.

\subsection{A family of $(\mu+\lambda)$ GA}
\label{sec:Algorithm}
To instantiate variants of the GA for the dynAS problem, we work on the configurable GA framework proposed in~\cite{Yeppsn2020}, which allows us to tune parameters and select from a set of operators. This framework can also be used for a future extension of this study to the dynAC problem. 
Algorithm~\ref{alg:GA} presents the details of the framework.

The GA initializes its population uniformly at random. For each iteration, $\lambda$ offspring is created either by using crossover (with probability $p_c$) or using mutation (with probability $1-p_c$), and the best $\mu$ of parent and offspring individuals are selected for the parent population of the next iteration. The GA terminates until hitting the optimum or reaching the maximal budget of function evaluations.

Three well-known crossover operators, \textit{one-point crossover}, \textit{two-point crossover}, and \textit{uniform crossover}, and two mutation operators, \textit{standard bit mutation} and \textit{fast mutation} are optional for the GA framework. For the \textit{standard bit mutation}, we flip bits at $\ell$ distinct positions, which are randomly chosen. $\ell$ is sampled from a conditional binomial distribution Bin$_{>0}(d,p)$~\cite{jansen2011analysis}, where $d$ is the dimension and $p$ is fixed as $1/d$ in this paper. For the \textit{fast mutation}, $\ell$ is sampled from a power-law distribution, and we follow the suggestion in~\cite{doerr2017fast}. 

\begin{algorithm2e}[htb]
\textbf{Input:} Population sizes $\mu$, $\lambda$, crossover probability $p_c$, mutation rate $p$\;
\textbf{Initialization:} 
    \lFor{$i=1,\ldots,\mu$}{sample $x^{(i)} \in \{0,1\}^d$ uniformly at random (u.a.r.), and evaluate $f(x^{(i)})$}
		Set $P = \{x^{(1)},x^{(2)},..., x^{(\mu)}\}$\;
	\textbf{Optimization:}
	\For{$t=1,2,3,\ldots$}{
	    $P' \leftarrow \emptyset$\;
	    \For{$i=1,\ldots,\lambda$}{
	        Sample $r \in [0,1]$ u.a.r.\;
	        \eIf{$r \le p_c$}{
	            select two individuals $x,y$ from $P$ u.a.r. (with replacement)\;
	            $z^{(i)} \assign \text{Crossover}(x,y)$\;
	            \lIf{$z^{(i)} \notin \{x,y\}$}{evaluate $f(z^{(i)})$ \\
	            \textbf{else} infer $f(z^{(i)})$ from parent}
	        }{
	            select an individual $x$ from $P$ u.a.r.\;
	            $z^{(i)} \assign \text{Mutation}(x)$\;
	            \lIf{$z^{(i)} \neq x$}{evaluate $f(z^{(i)})$ 
	            \\ \textbf{else} infer $f(z^{(i)})$ from parent}
	        }
	        $P'\leftarrow P'\cup\{z^{(i)}\}$\;
	    }
        $P$ is updated by the best $\mu$ points in $P \cup P'$ (ties broken u.a.r.)\;
	}
\textbf{Terminal Condition:} The optimum is found or the budget is used out\;
\caption{A Family of $(\mu+\lambda)$~Genetic Algorithms}
\label{alg:GA}
\end{algorithm2e}

\subsection{The IOHprofiler Problem Set}
\label{sec:Problems}
The IOHprofiler problem set~\cite{doerr2020benchmarking} initially contains 23 real-valued pseudo-Boolean problems, and another two problems were added in~\cite{Yeppsn2020}. Based on the prior data set of 80 variants of GAs, which is available at~\cite{FURONG2020_3752064}, we can investigate the potential improvement that could be theoretically obtained from applying the dynAS.

For the ease of understanding the following discussion, we provide partial definitions of the problems below. Details of the other problems are available in~\cite{doerr2020benchmarking,Yeppsn2020}.
\begin{itemize}
    \item \textbf{F1: \onemax} is maximizing the number of ones, which asks to maximize the function OM: $\{0,1\}^n \rightarrow [0..n], x \mapsto \sum x_i$.
    \item \textbf{F2: \leadingones} is maximizing the number of initial ones, which asks to maximize the function LO: $\{0,1\}^n \rightarrow [0..n]$, $x \mapsto \max\{i \in [0..n]|\forall j \le i: x_j = 1 \}$.
    \item \textbf{F3: A linear function with Harmonic Weights.} We can see this function as a variant of \onemax with weighted variables, which asks to maximize: $\{0,1\}^n \rightarrow \mathcal {R}, x \mapsto  \sum ix_i$.
    \item \textbf{F5: A W-model extension of \onemax (Reduction).} \\Dummy variables are introduced to \onemax for this function. $10\%$ randomly selected bits do not have any impact on the fitness value. Therefore, for a $n$-dimensional F5, its optimum is $0.9n$. For F4, $50\%$ bits do not have any impact on the fitness value, the others are the same.
    \item \textbf{F6: A W-model extension of \onemax (Neutrality).} The original input bit-string $(x_1,...,x_n)$ is mapped to a bit-string $(y_1,...,y_{\lfloor n/3 \rfloor})$. The value of $y_i$ is the majority of \\$(x_{3i - 2}, x_{3i-1}, x_{3i})$. The fitness value of $x$ on F6 is OM$(y)$. Therefore, for a $n$-dimensional F6, its optimum is $\lfloor n/3 \rfloor$.
    \item \textbf{F7:} \textbf{A W-model extension of \onemax (Epistasis)}. An epistasis function is applied to disturb permutation of bit-strings. Assuming there are two bit-strings $b_1,b_2$ with Hamming distance $1$, after the transformation with the epistasis function, the distance between two transformed bit-strings $b_1',b_2'$ is $\upsilon - 1$, where $\upsilon$ is the length of the bit-string. F7 partitions the input bit-string into segments of length $\upsilon = 4$ and applies the epistasis function on each segment. More details can be found in Sec 3.7.3 of~\cite{doerr2020benchmarking}.
    \item \textbf{F8:} \textbf{A W-model extension of \onemax (Ruggedness)}. Ruggedness is introduced by performing a transformation function on the fitness value OM$(x)$. The transformation function $r: [0..d] \rightarrow [0..\lceil d/2 \rceil +1]$ is defined as follows: $r(d) = \lceil d/2 \rceil +1$, $r(i)=\lfloor d/2 \rfloor + 1$ if $i$ is even and $i < d$, and $r(i)=\lceil d/2 \rceil + 1$ if $i$ is odd and $i < d$. The fitness value of F8 is $r(\text{OM}(x))$
    \item \textbf{F24: Concatenated Trap (CT)} partitions the input bit-string into segments of length $k$ and returns the sum of fitness values of concatenating Trap functions that take the segments as input. The Trap function asks to maximize Trap: $\{0,1\}^k \rightarrow [0,1]$. Trap$(x) = 1$ if the number $u$ of ones is equal to $k$, otherwise, Trap$(x)= (k-1-u)/k$. $k$ sets as $5$.
\end{itemize}

\section{Dynamic Crossover Probability Selection: A Study on \leadingones}\label{sec:LO}
Inspired by the investigation in~\cite{Yeppsn2020} that, on \leadingones, the optimal crossover probability of Algorithm~\ref{alg:GA} is dynamic along the problem dimension and population size, we are interested in the performance of the GAs with using uniform crossover at different stages.

\textbf{Dynamic optimal crossover probability.} To obtain the \textit{optimal} crossover probability at different stages of the algorithm, we test the $(10+10)$~GA using standard bit mutation with $p=1/n$ and uniform crossover with different $p_c \in \{0.1k \mid k \in [0..9]\}$. 
Algorithms run at stages of fitness value $f \in [s,s+5], s \in \{5i \mid i \in [0..19]\}$ on $100$-dimensional \leadingones. Practically, we initialize the population of the GAs with all the individual's fitness values equal to $s$, and the algorithms terminate once a solution with $f(x) \ge s+5$ is found. 

\begin{figure}[htb]
 \includegraphics[width=0.8\linewidth, trim = 0mm 5mm 0mm 0mm, clip]{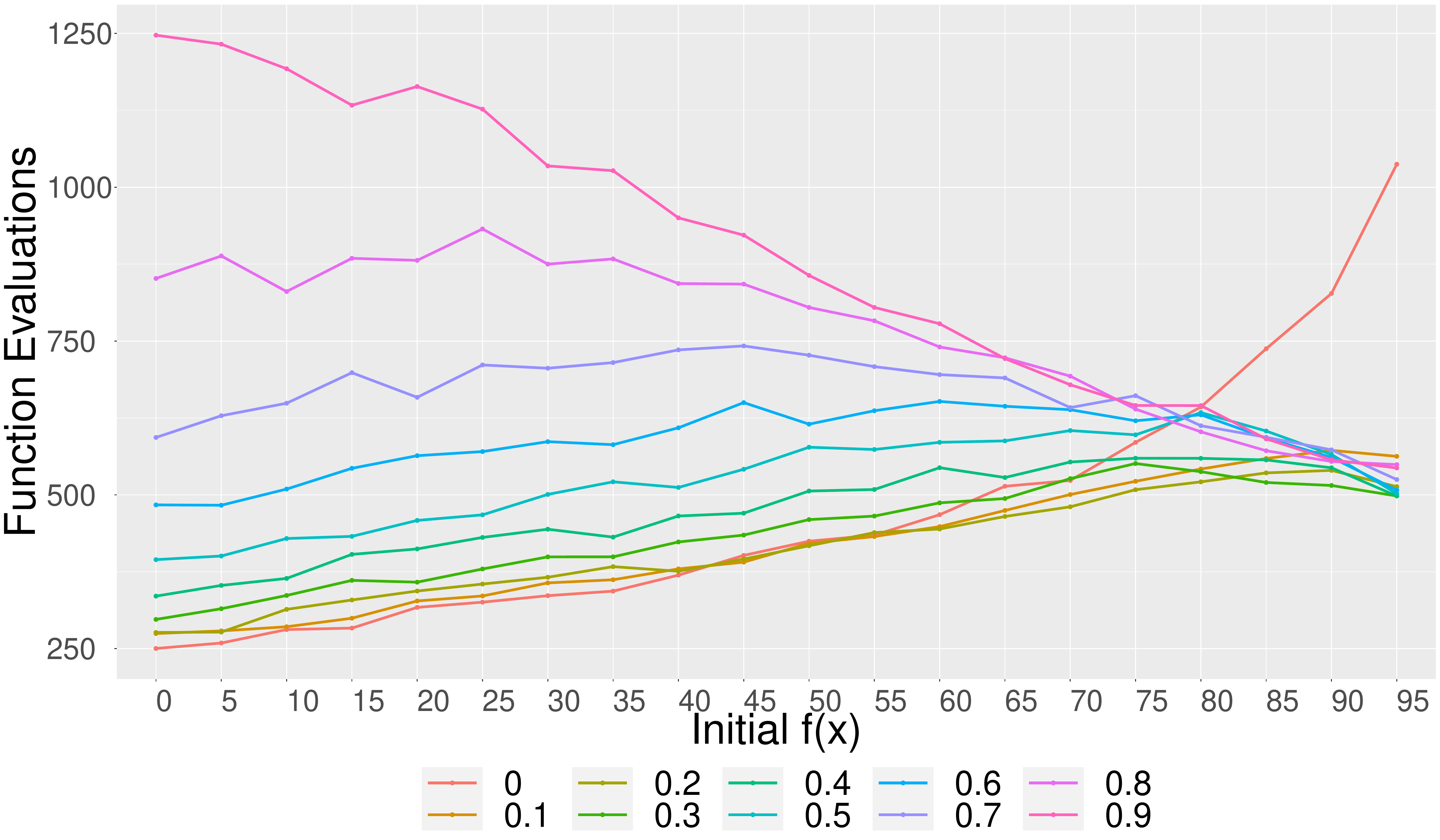}
  \vspace{-0.3cm}
\caption{Average number of function evaluations needed by different $(10+10)$~GAs to find a solution $y$ with $f(y) \geq s + 5$ on the $100$-dimensional \leadingones function when all ten points in the initial population are uniformly chosen from the set of points $x$ that  satisfy $f(x)=s$, for $s \in \{5i \mid i \in [0..19]\}$. The GAs differ only in the crossover probability $p_c \in \{0.1k \mid k \in [0..9]\}$ (different lines). Results are averaged of $1\,000$ independent runs. The connecting lines are only meant to help visual interpretation, the data points are only at the values 0, 5, 10, ..., 95. }
\vspace{-0.2cm}
 \label{fig:Seg-ERT-LO}
\end{figure}

Figure~\ref{fig:Seg-ERT-LO} plots function evaluations used by the GAs at each stage. It shows that the GA with $p_c=0$ spends the least function evaluations at the early stages $s \le 40$, but is outperformed by other GAs with $p_c>0$ as $s$ increasing. With the observation on the population of the GAs at late stages, we find that, for the GA with $p_c>0$, the fitness of most individuals converges quickly to the best found fitness after a better solution is found, but for the GA with $p_c = 0$, the fitness of most individuals remains constant.
when the best solution individual has been updated several times. An intuitive explanation for why the former performs better at later stages is that the GA can copy the current best initial ones to increase the quality of the whole population by using uniform crossover.

\textbf{Dynamic crossover probability selection.} With the result in Figure~\ref{fig:Seg-ERT-LO}, we expect to gain improvement by using the \textit{optimal} crossover probability at all stages. Figure~\ref{fig:Theo-dynGA} plots the fixed-target ERTs of GAs with static $p_c$ and dynamic ones. The dynamic policy selects the corresponding best $p_c$ at each stage. Practically, as the GA finds a solution with $s_1 \le f(x) < s_2, s_2 = s_1 + 5, s \in \{5i \mid i \in [0..19]\}$, the $p_c$ will adjust by using the corresponding best value in Figure~\ref{fig:Seg-ERT-LO}. In other words, the dynamic policy is a dynAS policy $\pi$ in which inputs are $P$ is \leadingones, $\mathcal{A}$ consists of $10$ GAs with different $p_c$, and $S$ is the set of $20$ targets $s$. 

\begin{figure}[htb]
 \includegraphics[width=0.9\linewidth, trim = 0mm 0mm 0mm 0mm, clip]{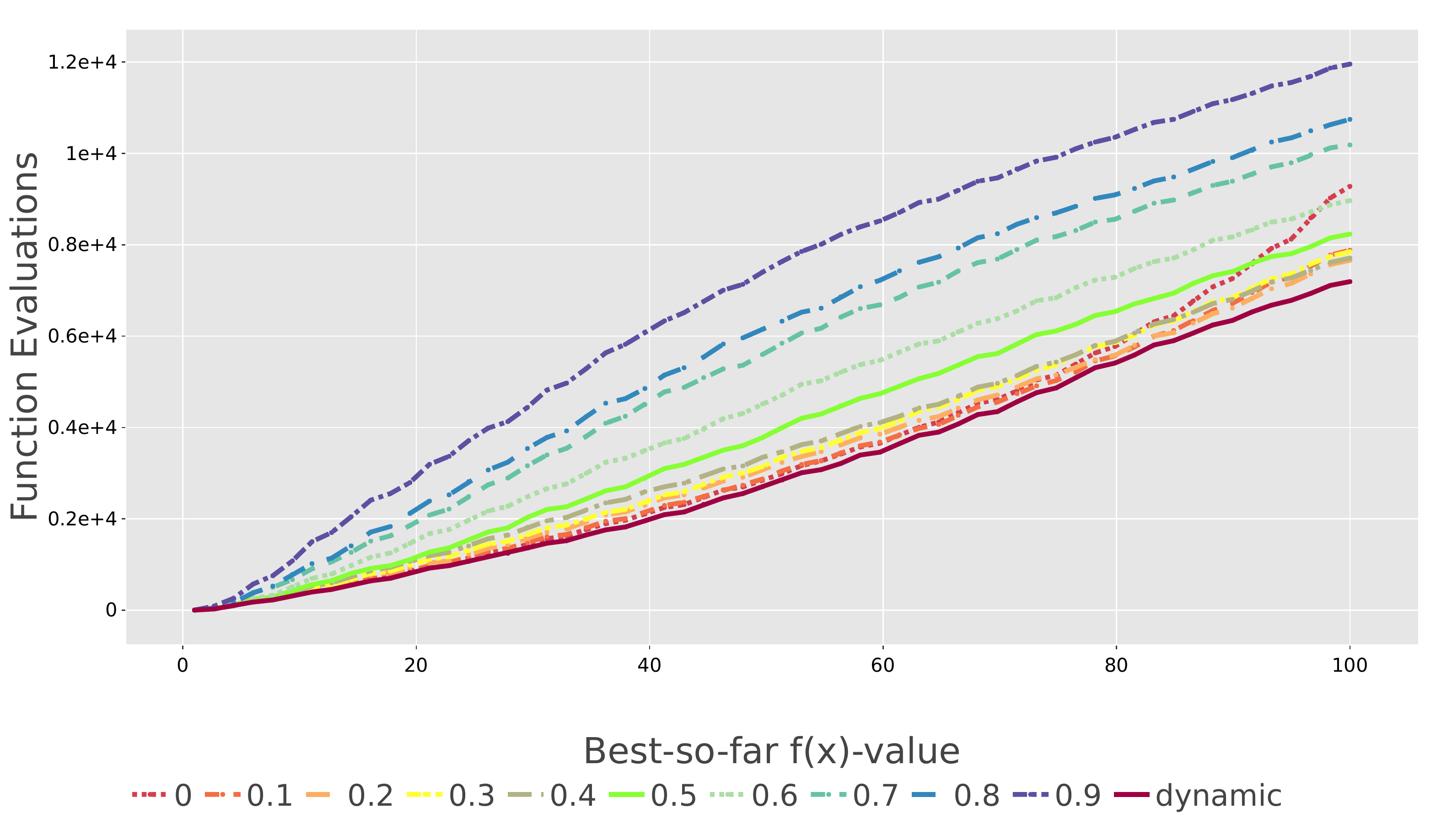}
  \vspace{-0.3cm}
\caption{Fixed-target ERTs of GAs on $100$-dimensional \leadingones. The legend presents values of $p_c$, and the \textit{dynamic} adjusts its $p_c$ to the \textit{optimal} value at each target $f(x)=s, s \in \{5i \mid i \in [0..19]\}$, based on the result in Figure~\ref{fig:Seg-ERT-LO}. Results are average of $100$ independent runs. The figure is produced using the IOHprofiler tool~\cite{doerr2018iohprofiler}. }
\vspace{-0.4cm}
 \label{fig:dynamicGA-LO}
\end{figure}

We observe in Figure~\ref{fig:dynamicGA-LO} that the GA with dynamic $p_c$ outperforms other GAs at all points in time, which leads to a success hitting the optimum $f(x)=100$ with the smallest ERT. 
Concretely, the ERT of the dynamic policy is $7\,194$, whereas that of the best runner-up (the GA with $p_c=0.2$) is $7\,661$. This corresponds to a $6\%$ improvement of the dynamic GA over the best static one. This performance empirically proves that the GA can benefit from dynamic crossover probability, and it displays a successful case of applying the dynAS for the GA. However, the dynAS problem is not usually coming with with the ideal condition that candidate algorithms differ by only one parameter, so that we are considering GAs with more combinations of parameters and operators in the next section.

\section{Dynamic Algorithm Section for the IOHprofiler problems}\label{sec:IOH}
\label{dynAS-PBO}
Since the \leadingones case shows significant improvement by using dynamic crossover probabilities, which is a particular case of the dynAS, we  study the behavior of the dynAS on a broader range of problems and GAs. In this section, we apply the dynAS on the 25 IOHprofiler benchmark problems (see Sec~\ref{sec:Problems}) with considering $80$ GAs. The optional parameter settings and operators for the GA (see Sec\ref{sec:Algorithm}) are listed below:
\begin{itemize}
    \item $4$ population size schemata: $(\lambda+1),(\lambda+\lambda/2)$, and $(\lambda + \lambda), \lambda \in \{10,50,100\}$, and $(1+\lambda), \lambda \in \{1,10,50,100\}$.
    \item $2$ mutation operators: \textit{standard bit mutation} (sbm) and \textit{fast mutation}.
    \item $3$ crossover operators: \textit{one-point crossover}, \textit{two-point crossover}, 
    and \textit{uniform crossover}.
    \item $2$ crossover probabilities: $p_c \in \{0,0.5\}$
\end{itemize}

Crossover operators are only applied for $(\lambda+1),(\lambda+\lambda/2)$, and $(\lambda + \lambda)$~ GAs with $p_c=0.5$, and $(1+\lambda)$~GAs are all mutation-only GAs.

\subsection{Theoretical Improvement}
\begin{figure}[htb]
 \includegraphics[width=0.95\linewidth, trim = 0mm 0mm 0mm 0mm, clip]{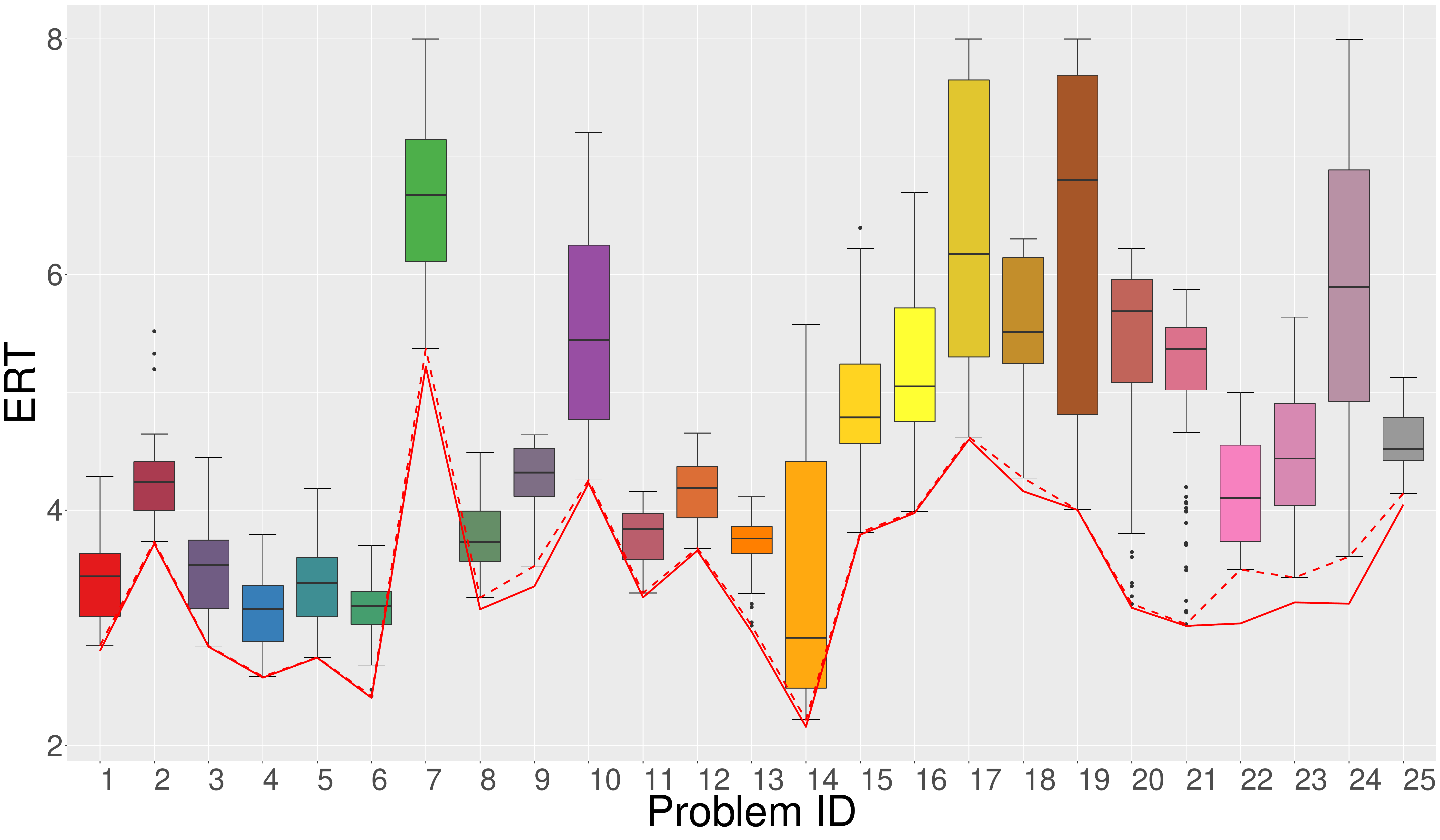}
  \vspace{-0.2cm}
\caption{Distributions of $\log_{10}$ERTs among all $80$ GAs on the 25 IOHprofiler problems in dimension $d=100$. The dashed line connects the points of the best ERTs for each problem. The solid line connects the predicted ERTs of the best dynGAs. Experimental results are from $100$ independent runs.}
\vspace{-0.5cm}
 \label{fig:Theo-dynGA}
\end{figure}

At first, we investigate the performance of the $80$ static GAs. Figure~\ref{fig:Theo-dynGA} shows the distributions of ERTs among 25 functions. The targets used to calculate ERTs are listed in Table~\ref{tab:TheoResult}. We observe substantial differences among algorithms as well as among problems. A red dashed line connects the best ERTs of static GAs on each problem.

Based on the data in~\cite{FURONG2020_3752064}, we can calculate theoretical performance (predicted ERTs in formula~\ref{formula:pre-ERT}) of all possible policies $\pi$ with combinations of $80$ GAs.  As mentioned in Sec~\ref{sec:ERT-dynAS}, we consider the \textit{switch-once} dynAS. To generate the set $\Phi$ of targets, we select $19$ evenly spaced partition points within $[\phi_m,\phi_f]$ by linear scale and log scale respectively, where $\phi_m$ is the smallest fitness value of the problem, and $\phi_f$ is the final target. Note that we only consider the GAs that hit the corresponding target with a success rate $ps \ge 0.8$ for the dynAS.

Table~\ref{tab:TheoResult} lists the best dynamic algorithm policy (BDA, see \textit{Definition 2.2.2}) for the IOHprofiler problems, and their predicted ERTs are also visualized by a solid red line in Figure~\ref{fig:Theo-dynGA}. For ease of notation, we denote dynGA as the method of the dynAS policy. We expect the dynGA, which is the theoretically best, to outperform the BSA on all $25$ problems. We also observe that the BSAs are usually selected for either the first or the second stage for the BDAs, expect for F7, F14, and F22-23. 
For the targets where the BDAs switch from using one algorithm to another one, they either are close to the final target or locate at the early stage, expect for F18 and F24.

\begin{table*}[ht]
\small
\centering
\begin{tabular}{rrrrrrrrrr}
   \hline
 funcId & fTarget & BSA & sERT & A1 & A2 & sTarget & dERT & ratio ($\%$) \\ 
    \hline
1 & 100 & (1+1) EA$_{>0}$ & 705 & (1+1) EA$_{>0}$ & (10+10)-uniform-GA & 96 & 638 & 9.5 \\ 
2 & 100 & (1+1) EA$_{>0}$ & 5\,430 & (1+1) fast GA & (1+1) EA$_{>0}$ & 21 & 5\,186 & 4.5 \\ 
3 & 5\,050 & (1+1) EA$_{>0}$ & 702 & (100+1)-two-point-fGA & (1+1) EA$_{>0}$ & 2\,899 & 693 & 1.3 \\ 
4 & 50 & (1+1) EA$_{>0}$ & 387 & (1+1) EA$_{>0}$ & (1+10) EA$_{>0}$ & 49 & 379 & 2.1 \\ 
5 & 90 & (1+1) EA$_{>0}$ & 562 & (1+1) fast GA & (1+1) EA$_{>0}$ & 55 & 559 & 0.5\\ 
6 & 33 & (1+1) EA$_{>0}$ & 265 & (100+100) EA$_{>0}$ & (1+1) EA$_{>0}$ & 20 & 255 & 3.8 \\ 
7 & 100 & (50+50) EA$_{>0}$ & 234\,980 & (100+100)-two-point-GA & (100+50) EA$_{>0}$ & 95 & 182\,271 & 22.4 \\ 
8 & 51 & (10+10)-uniform-GA & 1\,808 & (10+10)-uniform-GA & (50+50)-uniform-GA & 50 & 1\,441 & 20.3 \\ 
9 & 100 & (100+1)-uniform-fGA & 3\,354 & (50+1)-uniform-fGA & (100+100)-uniform-fGA & 96 & 2\,255 & 32.8 \\ 
10 & 100 & (100+100)-uniform-fGA & 53\,083 & (50+25)-uniform-fGA & (100+100)-uniform-fGA & 94 & 16\,956 & 68.1 \\ 
11 & 50 & (1+1) EA$_{>0}$ & 1\,982 & (1+1) fast GA & (1+1) EA$_{>0}$ & 17 & 1\,818 & 8.3 \\ 
12 & 90 & (1+1) EA$_{>0}$ & 4\,764 & (1+1) fast GA & (1+1) EA$_{>0}$ & 19 & 4\,535 & 4.8 \\ 
13 & 33 & (1+1) EA$_{>0}$ & 1\,047 & (1+1) fast GA & (1+1) EA$_{>0}$ & 14 & 929 & 11.3 \\ 
14 & 7 & (100+1)-uniform-fGA & 166 & (100+50)-one-point-GA & (50+25)-uniform-fGA & 5 & 145 & 12.7 \\ 
15 & 51 & (1+1) EA$_{>0}$ & 6\,474 & (1+1) fast GA & (1+1) EA$_{>0}$ & 10 & 6\,200 & 4.2 \\ 
16 & 100 & (1+1) EA$_{>0}$ & 9\,768 & (1+1) fast GA & (1+1) EA$_{>0}$ & 21 & 9\,455 & 3.2 \\ 
17 & 100 & (1+1) EA$_{>0}$ & 41\,697 & (1+1) fast GA & (1+1) EA$_{>0}$ & 21 & 40\,350 & 3.2 \\ 
18 & 4.22 & (50+50) fast GA & 240\,145 & (10+10) EA$_{>0}$ & (50+50) fast GA & 3.57 & 14\,468 & 94.0 \\ 
19 & 98 & (1+1) EA$_{>0}$ & 10\,048 & (1+1) fast GA & (1+1) EA$_{>0}$ & 60 & 10\,044 & 0.0 \\ 
20 & 180 & (1+1) EA$_{>0}$ & 1\,600 & (1+1) EA$_{>0}$ & (1+10) EA$_{>0}$ & 178 & 1\,482 & 7.4 \\ 
21 & 260 & (1+1) EA$_{>0}$ & 1\,076 & (1+1) EA$_{>0}$ & (1+10) EA$_{>0}$ & 258 & 1\,041 & 3.3\\ 
22 & 42 & (10+5)-two-point-fGA & 31\,920 & (1+10) EA$_{>0}$ & (100+1)-two-point-GA & 39 & 1\,092 & 96.6 \\ 
23 & 9 & (1+10) EA$_{>0}$ & 2\,682 & (1+1) EA$_{>0}$ & (10+1) EA$_{>0}$ & 8 & 1\,648 & 38.6 \\ 
24 & 17.20 & (100+100)-two-point-fGA & 4\,030 & (1+1) EA$_{>0}$ & (100+100)-two-point-fGA & 15.81 & 1\,607 & 60.1 \\ 
25 & -0.30 & (100+100)-uniform-fGA & 21\,208 & (1+1) fast GA & (100+100)-uniform-fGA & -0.32 & 11\,151 & 47.4 \\ 
   \hline
\end{tabular}
\caption{Theoretical Performance of the DAS for the 25 IOHprofiler benchmark problems in dimension $d=100$. $fTarget$ lists the final targets used to calculate ERTs, and $sERT$ lists the ERTs of the algorithms in column $staticAlg$, which are the best ones among the 80 tested GAs for each problem. The DAS switches from using $A1$ to using $A2$ as finding a solution with $f(x) \ge sTarget$, and the corresponding predicted ERTs are listed in $dERT$. $ratio=(sERT-dERT)/sERT$. All algorithms are tested with 100 independent runs.\protect\\
For algorithm names: 'EA$_{>0}$' denotes the mutation-only GAs using sbm with $p=1/n$, 'fast GA' denotes the mutation-only GAs using fast mutation. The GAs with $p_c=0.5$ are named as '$(\mu+\lambda)$-crossover operator-GA/fGA', where 'GA' indicates using sbm with $p=1/n$ and 'fGA' indicates using fast mutation.}
\vspace{-0.8cm}
\label{tab:TheoResult}
\end{table*}

\begin{figure}[htb]
 \includegraphics[width=0.95\linewidth, trim = 0mm 0mm 0mm 0mm, clip]{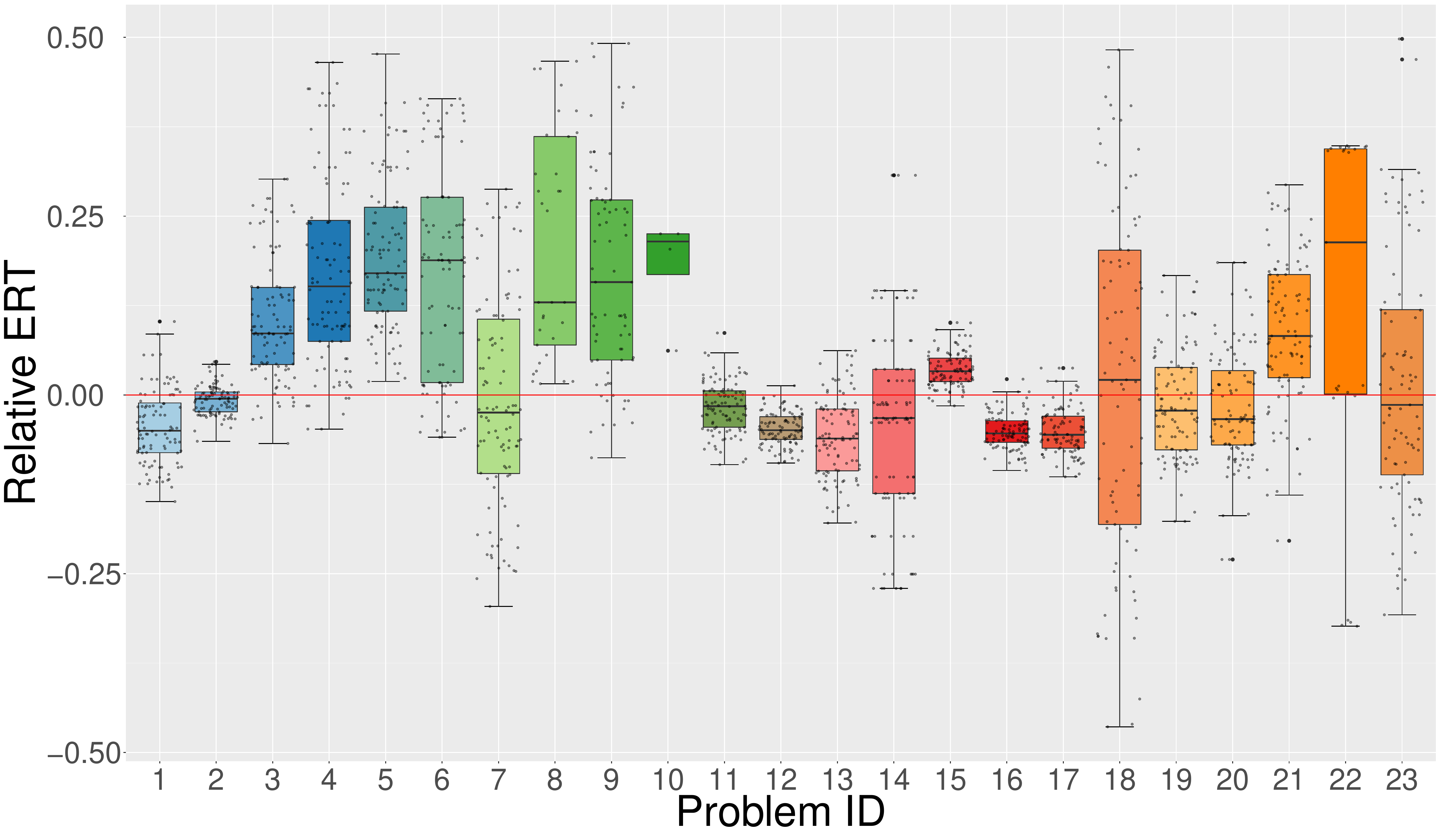}
  \vspace{-0.2cm}
 \caption{Box plots of relative ERTs of $100$ dynGAs ($dERT$) for the IOHprofiler problems in dimension $d=100$, comparing to the $sERT$ in Table~\ref{tab:TheoResult}. The relative deviation is calculated by $(dERT - sERT) / sERT$. Results of each algorithm are plotted by black dots. Negative values (below the red line) indicate better solvers comparing to the BSA. Values are capped by $[-0.5,-0.5]$ for visualization so that the results of F24-F25 are missing here with values larger than $1$.  Results are from 100 independent runs. Detailed data can be found at~\cite{furong_ye_2021_4501275blind}}
 \vspace{-0.5cm}
 \label{fig:Best100}
\end{figure}

\subsection{Experimental Result}
To reveal the practical performance of the predicted BDA and to study the behavior of the dynAS, instead of considering only the theoretically best one, we test $100$ dynGAs for each problem. Practically, we calculate the predicted ERTs of all combinations of $\pi$ over 80 algorithms and 42 targets and take the best $100$  for the experiment. For the dynGAs which the parent population sizes of $A_1$ and $A_2$ remain the same, we only adjust the parameter settings and operators as switching, for the dynGAs with $\mu_1 > \mu_2$, we selected the best $\mu_2$ of $\mu_1$ for the new parents after switching, and for the dynGAs with $\mu_1 < \mu_2$, the new parent population consist of copies of previous $\mu_1$, $\lfloor \mu_2 / 2 \rfloor - \mu_1$ copies of the best of previous $\mu_1$, and $\lceil \mu_2 / 2 \rceil$ new individuals randomly generated. The summary data of this paper can be found at \cite{furong_ye_2021_4501275blind}.

Figure~\ref{fig:Best100} plots the distributions of relative ERTs comparing to the $sERT$ in Table~\ref{tab:TheoResult}, and the result of each dynGA are marked by black dots. Note that we do not expect the entire group of $100$ dynGAs to perform better than the BSA because not all of the dynGAs can theoretically obtain ERTs better than $sERT$. However, as long as some dynGAs outperform the BSA (dots below the red line in Figure~\ref{fig:Best100}), we can expect an improvement by applying the dynAS.

We observe promising results of better solvers for problems (except F5, F8, F10, and F24-25) 
in Figure~\ref{fig:Best100}. On the other hand, we would like to investigate and better understand the unsuccessful trials.

\begin{figure}[!htb]
\centering
\subfigure[F3]{
\begin{minipage}{8cm}
\centering
\includegraphics[width=0.4\textwidth]{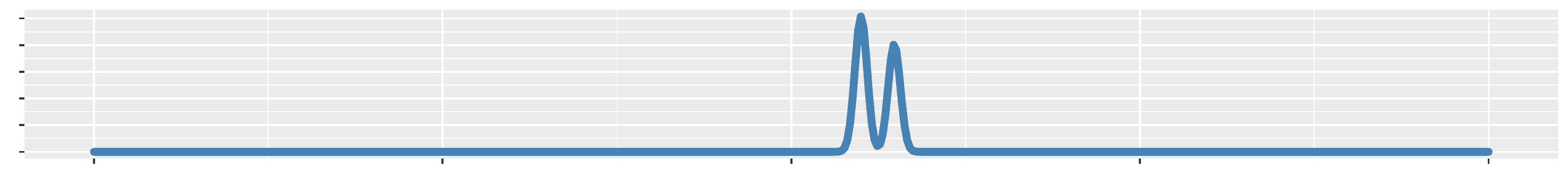}
\includegraphics[width=0.75\textwidth]{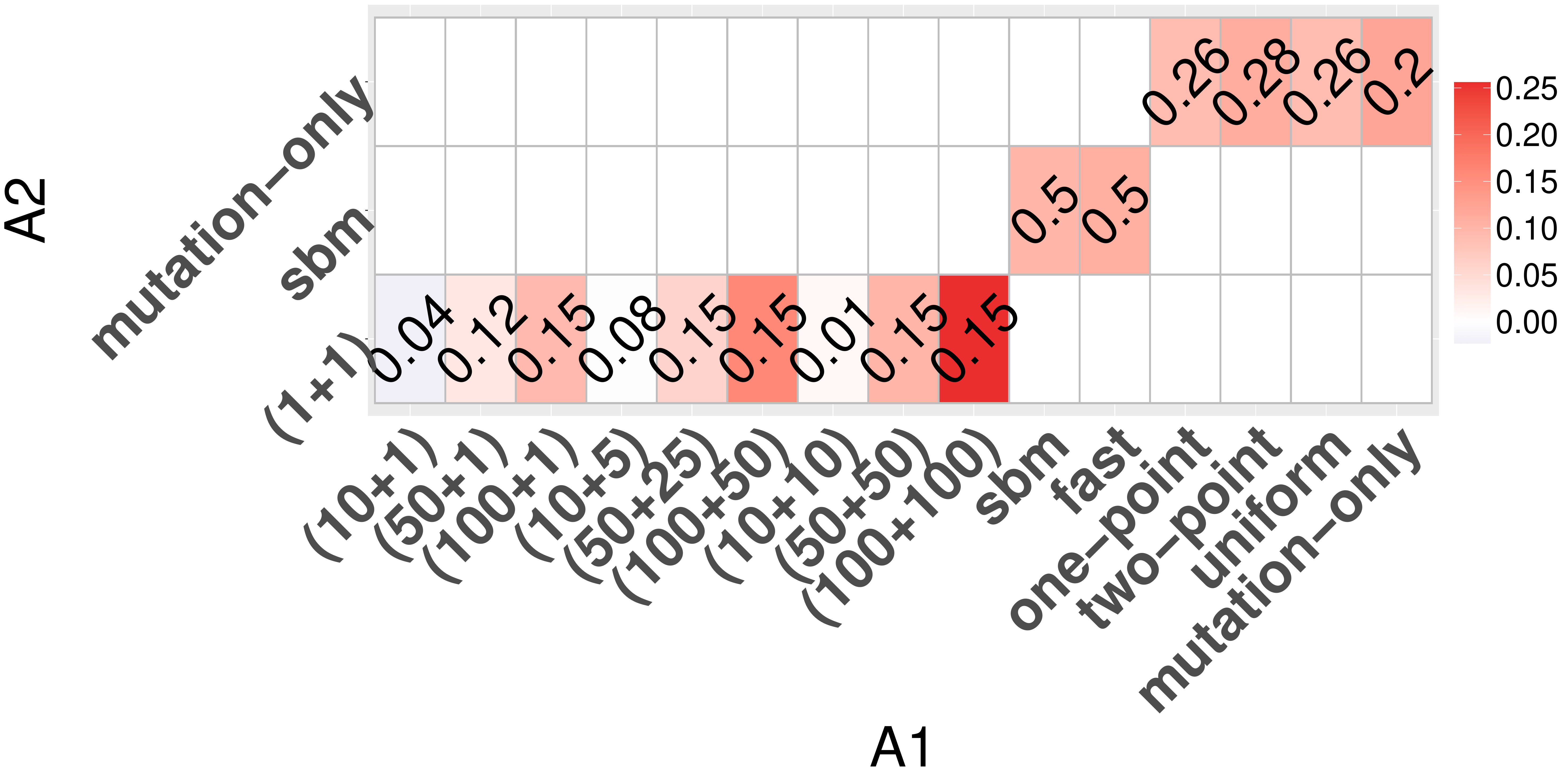}
\vspace{-0.5cm}
\end{minipage}
}

\subfigure[F5]{
\begin{minipage}{8cm}
\centering
\includegraphics[width=0.4\textwidth]{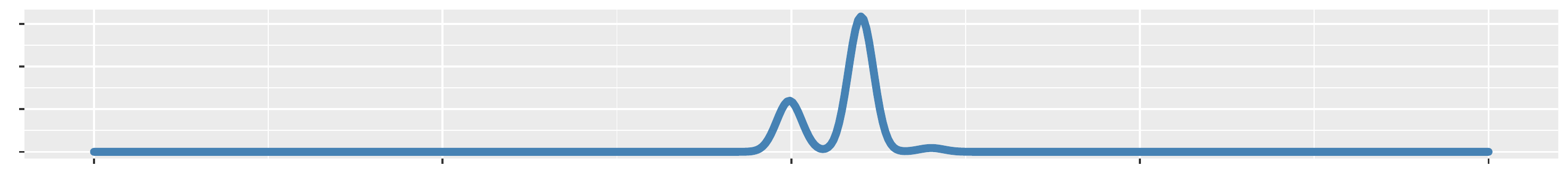}
\includegraphics[width=0.75\textwidth]{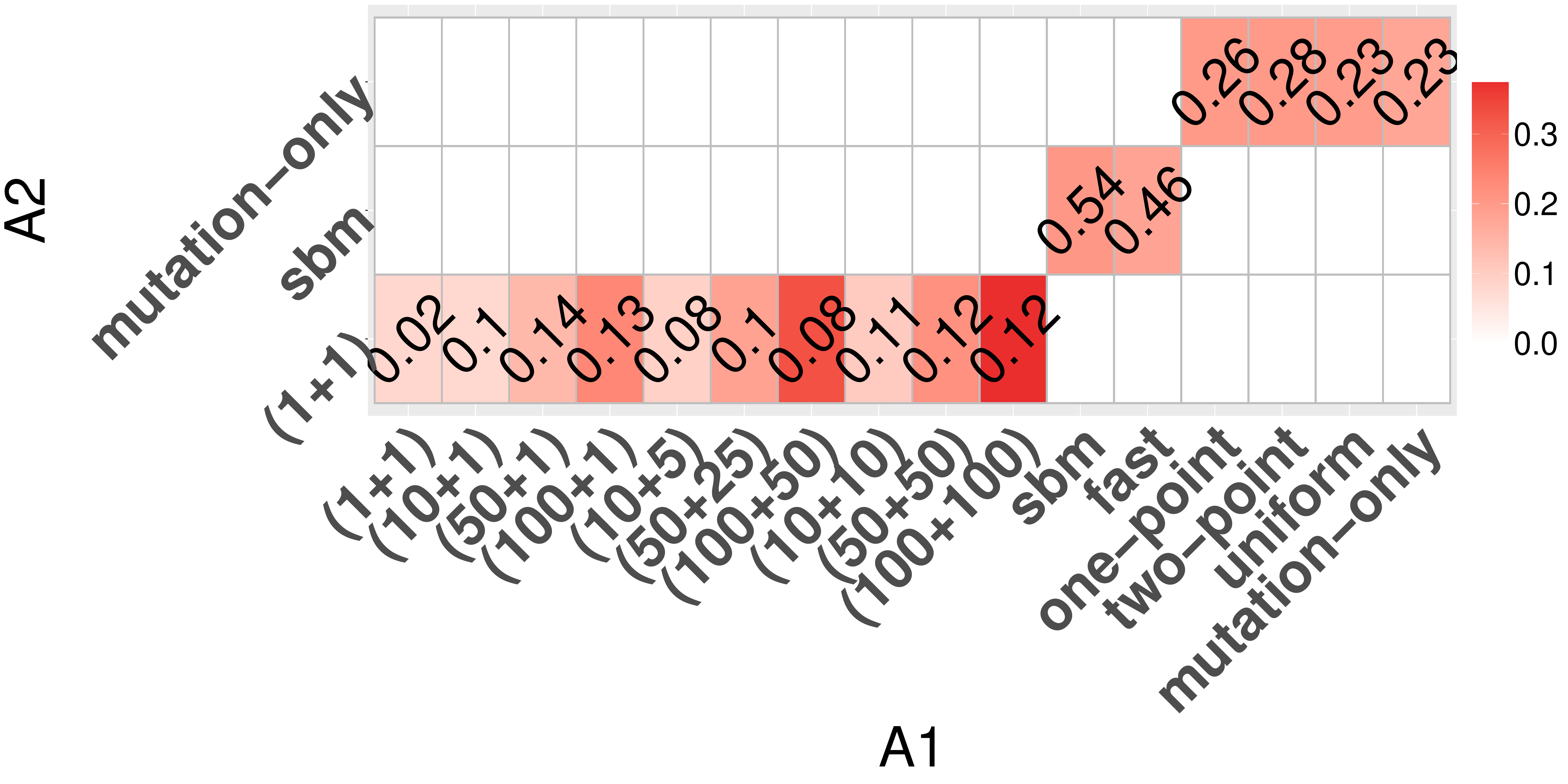}
\vspace{-0.5cm}
\end{minipage}

}
\subfigure[F6]{
\begin{minipage}{8cm}
\centering
\includegraphics[width=0.4\textwidth]{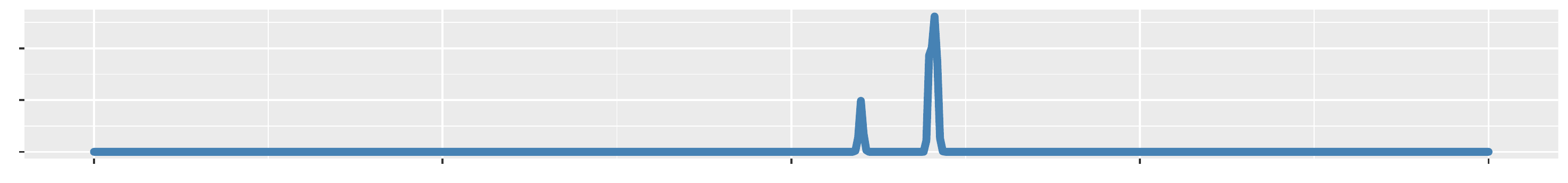}
\includegraphics[width=0.75\textwidth]{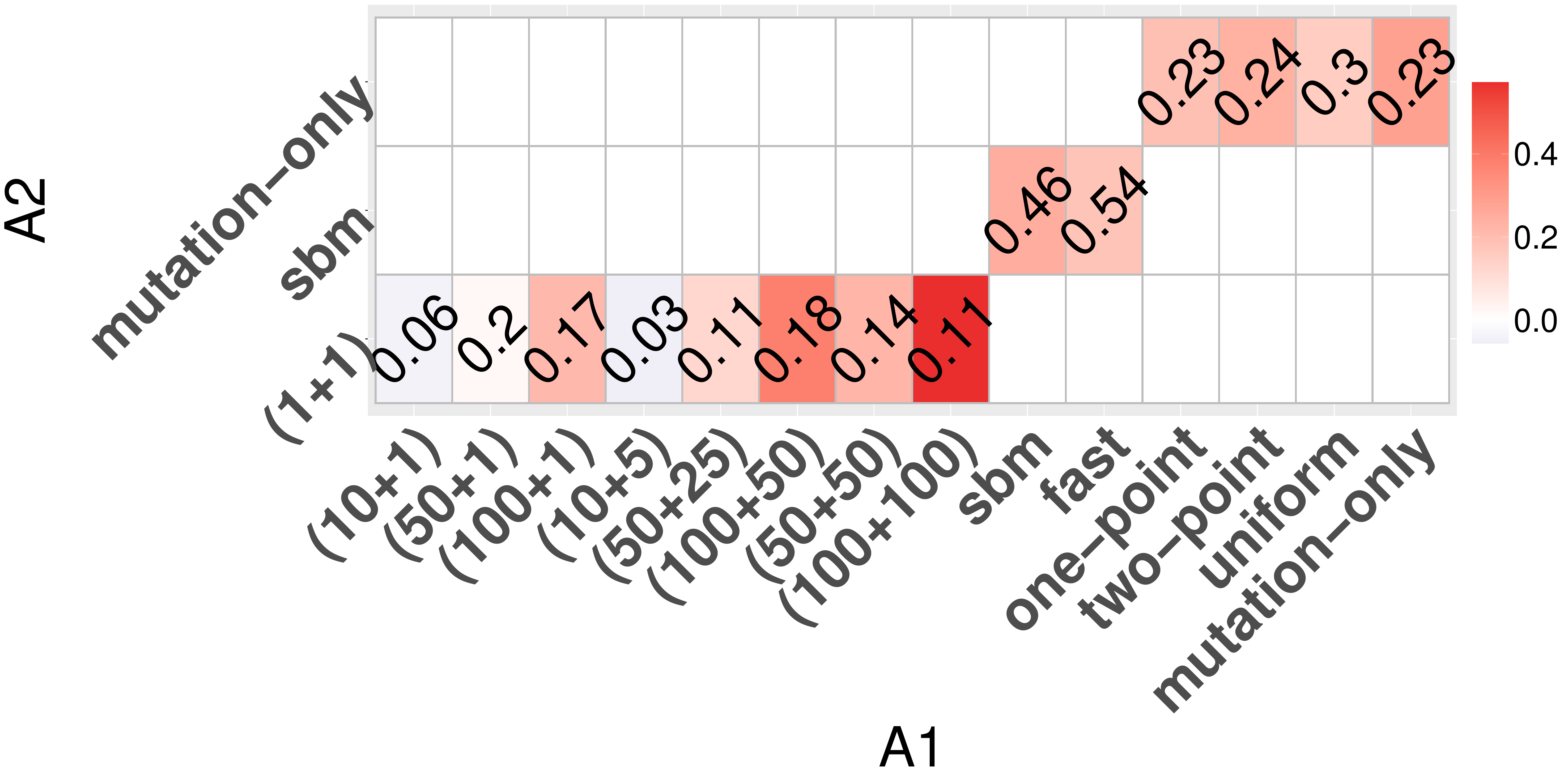}
\end{minipage}
}
\vspace{-0.2cm}
\caption{Averaged Relative ERTs of dynGAs with corresponding operator combination relative to the $sERT$ on F3-4, and F6 in dimension $d=100$. X-axis and Y-axis indicate the operators selected by $A_1$ and $A_2$ respectively. Purple color indicates better solvers comparing to the BSA. Numbers on tiles show the frequency of the combination that appears among $100$ algorithms.\\
The figures on the top of the tile plots are the distributions of $\phi_s$ of $100$ dynGAs, and values are scaled by $(\phi_s-\phi_m)/(\phi_f-\phi_m)$. $\phi_m$ is the minimal fitness of the problem.}
\vspace{-0.2cm}
\label{fig:F3-6Opt}
\end{figure}

Recall that a dynAS policy is described by $\pi=(A_1,A_2,\phi_s)$ with components of $A_1, A_2 \in \mathcal{A}$ and $\phi \in \Phi$. We at first discuss the experiment of F3-6 here concerning the limitation of candidate algorithms $\mathcal{A}$. Figure~\ref{fig:F3-6Opt} plots the frequencies of tested parameters and operators and the averaged relative ERTs of the dynGAs with the corresponding GAs combination. The frequency stands for theoretical prediction, and the relative ERT stands for the experimental result. Tiles are distributed at three zones in the figure: the bottom left is combinations of $(\mu+\lambda)$, the middle is combinations of mutation operators, and the upper right is combinations of crossover operators. We have erased the operators not being selected from the figures. For example, $0.04$ in the bottom left indicates $4$ dynGAs using $(10+1)$~GAs as the first algorithm and using $(1+1)$~GAs as the second one. The purple color indicates the averaged relative ERTs of these four dynGAs are less than 0 compared to the $sERT$.

For F3 and F5-6, the dynAS always chooses the \oea for A2, and it does not recognize mutation and crossover operators with different (dis)advantages for $A_1$. 
Looking at the $\phi_s$, we observe that switching is located around the initial fitness of these variants of \onemax problems. 
According to Table~\ref{tab:TheoResult}, the \oea performs  best on \onemax variants (F3-F6), and the dynAS is expected to gain improvement by switching at the very beginning. However, this may even not happen in practice because of the randomness of initialization. Also, due to the $sERT$ being relatively small on the problems, we can see that GAs using large $\mu$ deteriorate because it takes unnecessary evaluations for a large population. Uniform crossover has shown its advantages on \onemax in previous study~\cite{DoerrD18}, and we gain improvement by switching to use a GA with uniform crossover at late stages. However, for the \onemax variants with weighted variables, dummy variables, and neutrality, we do not observe that the dynGAs can benefit from uniform crossover.

\subsubsection{Diversity of the dynAS Policy}
\begin{figure}[htb]
\centering
\subfigure[Rank-select]{
\begin{minipage}{4cm}
\centering
\includegraphics[width=\textwidth]{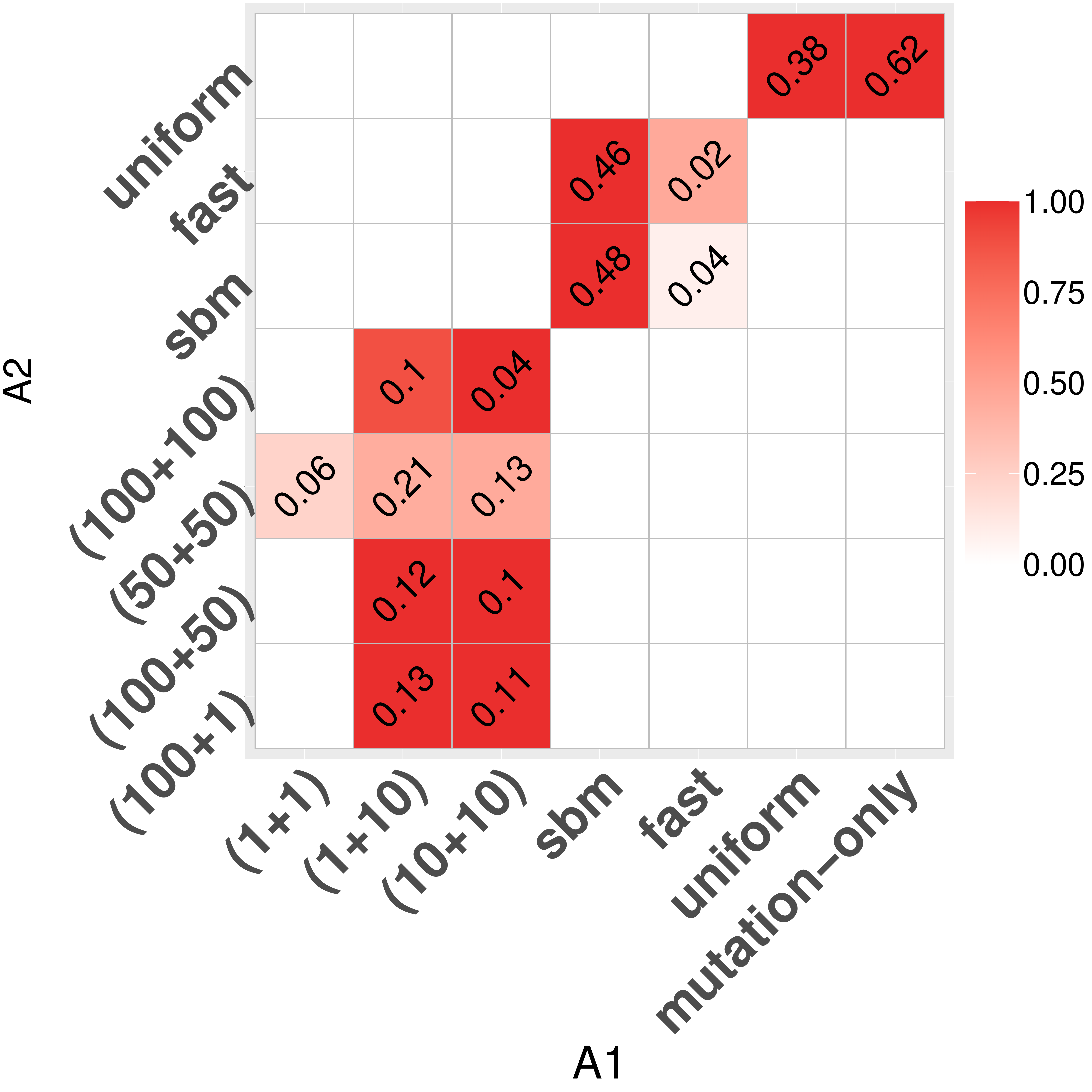}
\end{minipage}
}
\subfigure[Restriction-select]{
\begin{minipage}{4cm}
\centering
\includegraphics[width=\textwidth]{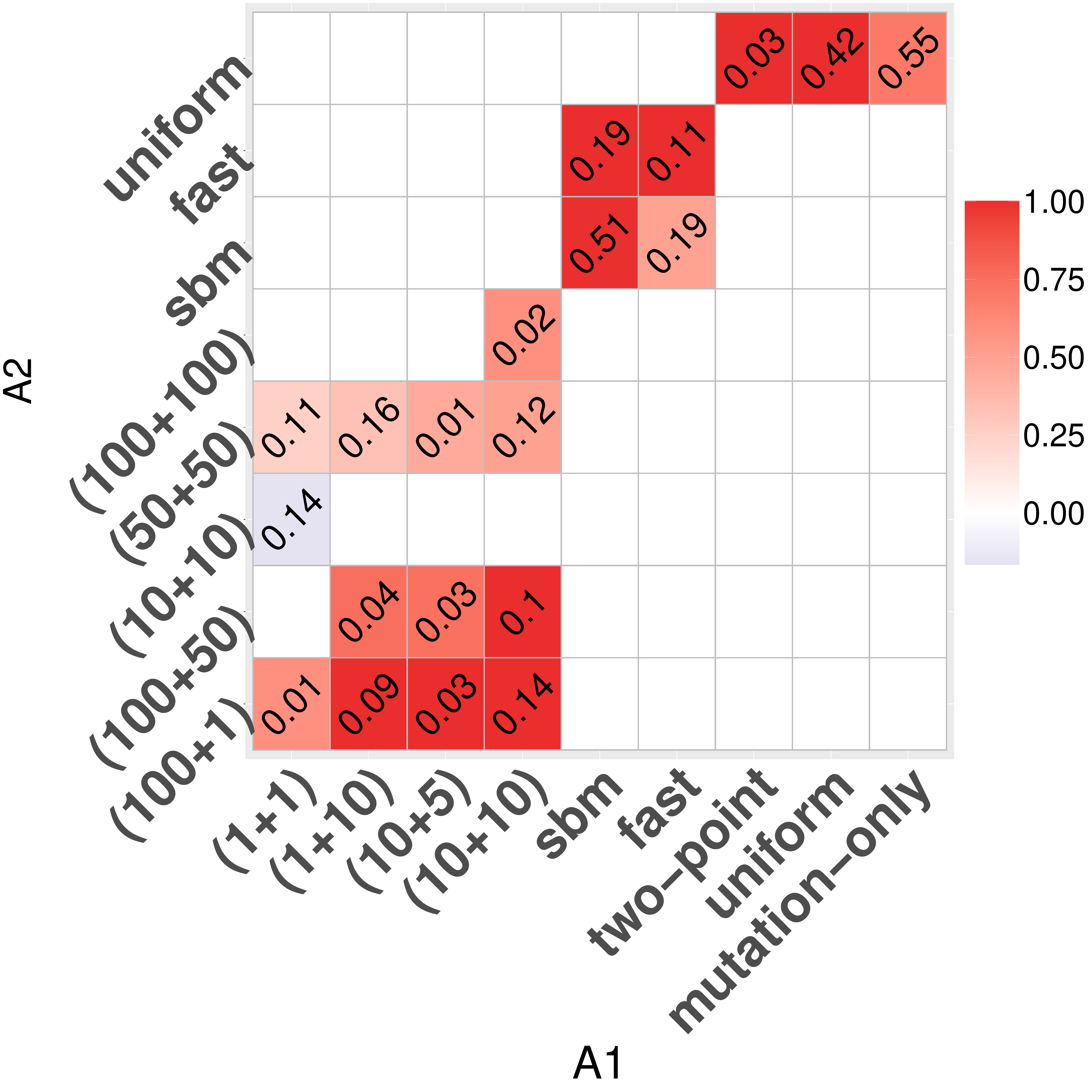}
\end{minipage}
}
\vspace{-0.5cm}
\caption{Averaged Relative ERTs of dynGAs with corresponding operator combination relative to the $sERT$ on F8 in dimension $d=100$. X-axis and Y-axis indicate the operators selected by $A_1$ and $A_2$ respectively. Purple color indicates better solvers comparing to the BSA. Numbers on tiles show the frequency of the combination that appears among $100$ algorithms.\\
The left figure plots the result of 100 dynGAs, which are the best 100 ranked by theoretical performance, and the selected times of the algorithms are capped by $20$ for the right figure.\\
 }
 \vspace{-0.2cm}
 \label{fig:F8-Operators}
\end{figure}

Apart from F3-6, we did not see significant improvement by using the dynAS for the \onemax variants F8 and F10. Recall that we study the informed dynAS so that we can obtain some preliminary information. Differently from the performance on F3-6, \oea is not the BSA for F8-10. The situations for F8 and F10 are similar, and F8 is taken here for the discussion. According to Figure~\ref{fig:F8-Operators}(a), the BSA $(10+10)$-uniform-GA is not selected for the tested dynGAs. $A_2$ of all dynGAs are still GAs using uniform crossover, but the parent population size $\mu > 10$. To explore potential improvement by increasing the diversity of the dynAS policy $\pi$, the number of a GA can not exceed $20$ when we select algorithms for $A_1$ and $A_2$, respectively. Figure~\ref{fig:F8-Operators}(b) plots the results of the dynGAs selected with such constraints. We observe that $(10+10)$~GAs are included for $A_2$, and the combination of using $(1+1)$~GAs as $A_1$ and using $(10+10)$~GAs as $A_2$ shows the only tile where improvement is obtained by average.

\begin{figure}[!htb]
\includegraphics[width=0.12\linewidth, trim = 0mm 0mm 0mm 0mm, clip]{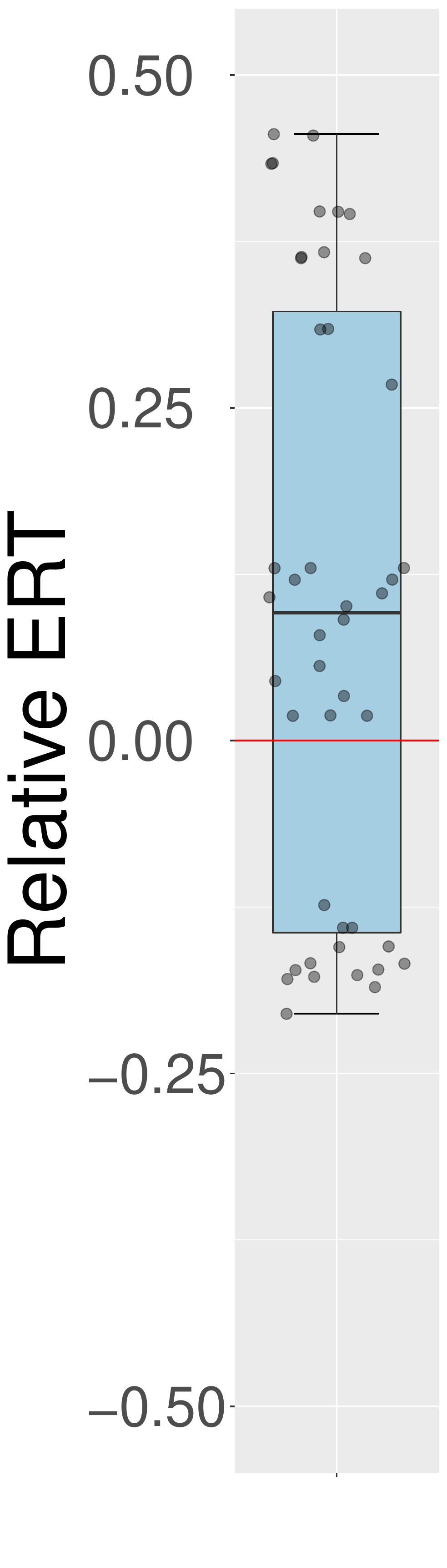} \quad \includegraphics[width=0.75\linewidth, trim = 0mm 2mm 0mm 0mm, clip]{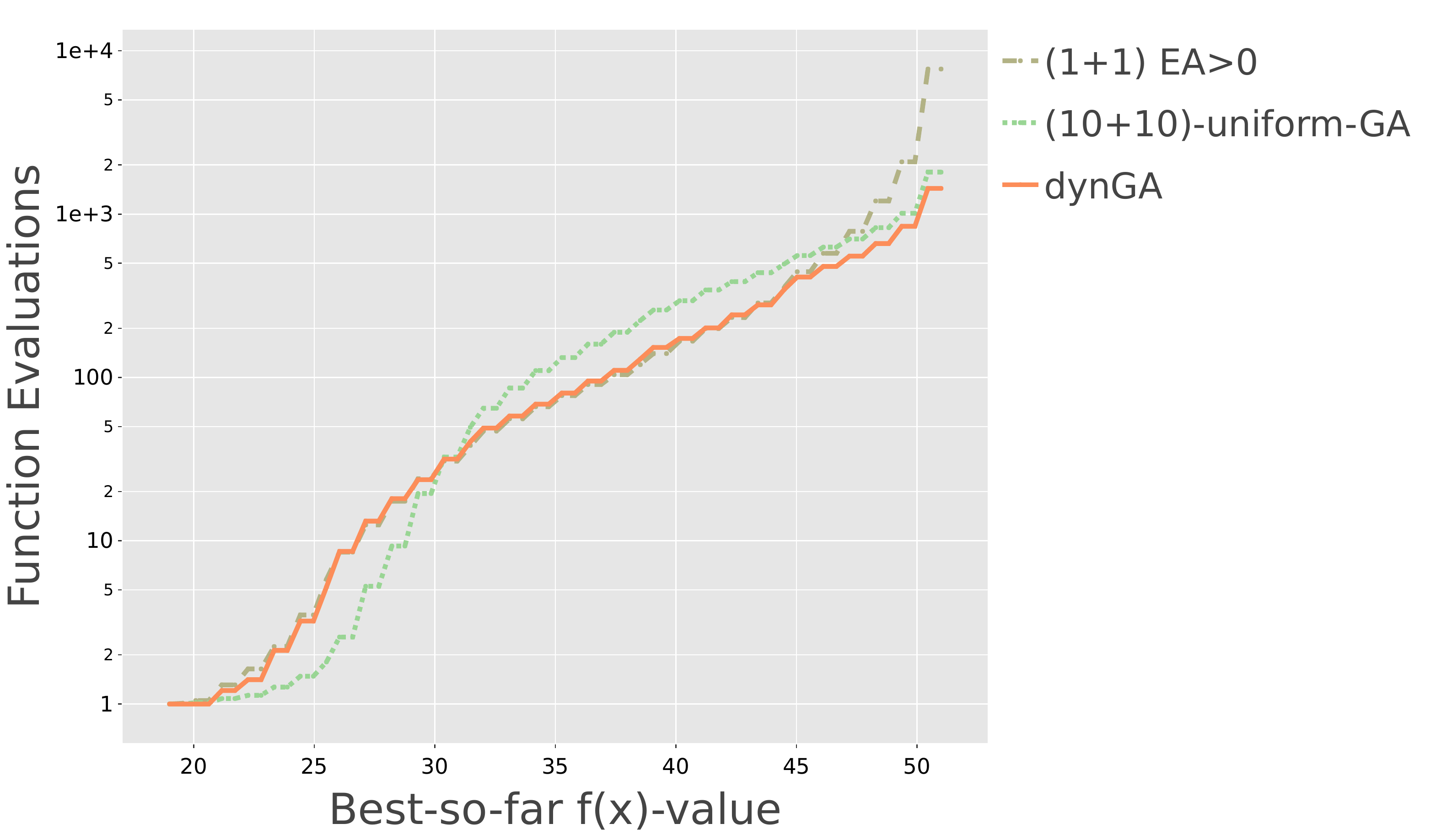}
 \vspace{-0.2cm}
\caption{The left is the box plot of relative ERTs comparing to the $sERT$ on F8 in dimension $d=100$. The right plots fixed-target ERTs of GAs. The dynGA switches from the $(10+10)$-uniform-GA to the $(1+1)$~EA$_{>0}$ at the target $f(x)=45$, which is produced using the IOHprofiler tool~\cite{doerr2018iohprofiler}.  Results are from 100 independent runs.}
\vspace{-0.5cm}
\label{fig:F8}
\end{figure}

Moreover, we plot the distribution of relative ERTs of the dynGAs selected with the constraints in Figure~\ref{fig:F8}, and the fixed-target result of the best one is given beside. The dynGAs with $(10+10)$~GAs as $A_2$ contribute all better solvers (dots below the red line). According to the fixed-target result, the dynGA benefits from uniform crossover at the late stage on F7, using fewer evaluation functions to handle the ruggedness and deceptiveness. Specifically, the dynGA using uniform crossover requires proper settings of $\mu$, according to Figure~\ref{fig:F8-Operators}, the advantage disappears as using $\mu = \{50,100\}$.

\subsubsection{Local Optima are Deceptive}
\label{sec:localoptima}
It is known that local optima bring difficulties for optimization, and in this work, we also observe the obstacle they cause for the dynAS. Recall that in formula~\ref{formula:pre-ERT} the contribution of $A_1$ to the predicted ERT is decided by its ERT hitting the target $f(x) = \phi_s$. However, using the ERT as the cost metric of the dynAS, we do not obtain more information to estimate if $A_1$ is trapped or around a local optimum.  
This lack of knowledge may affect the dynAS, and we observe it results in failures of this strategy for F24-25.

\begin{figure}[htb]
 \includegraphics[width=0.75\linewidth, trim = 0mm 0mm 0mm 0mm, clip]{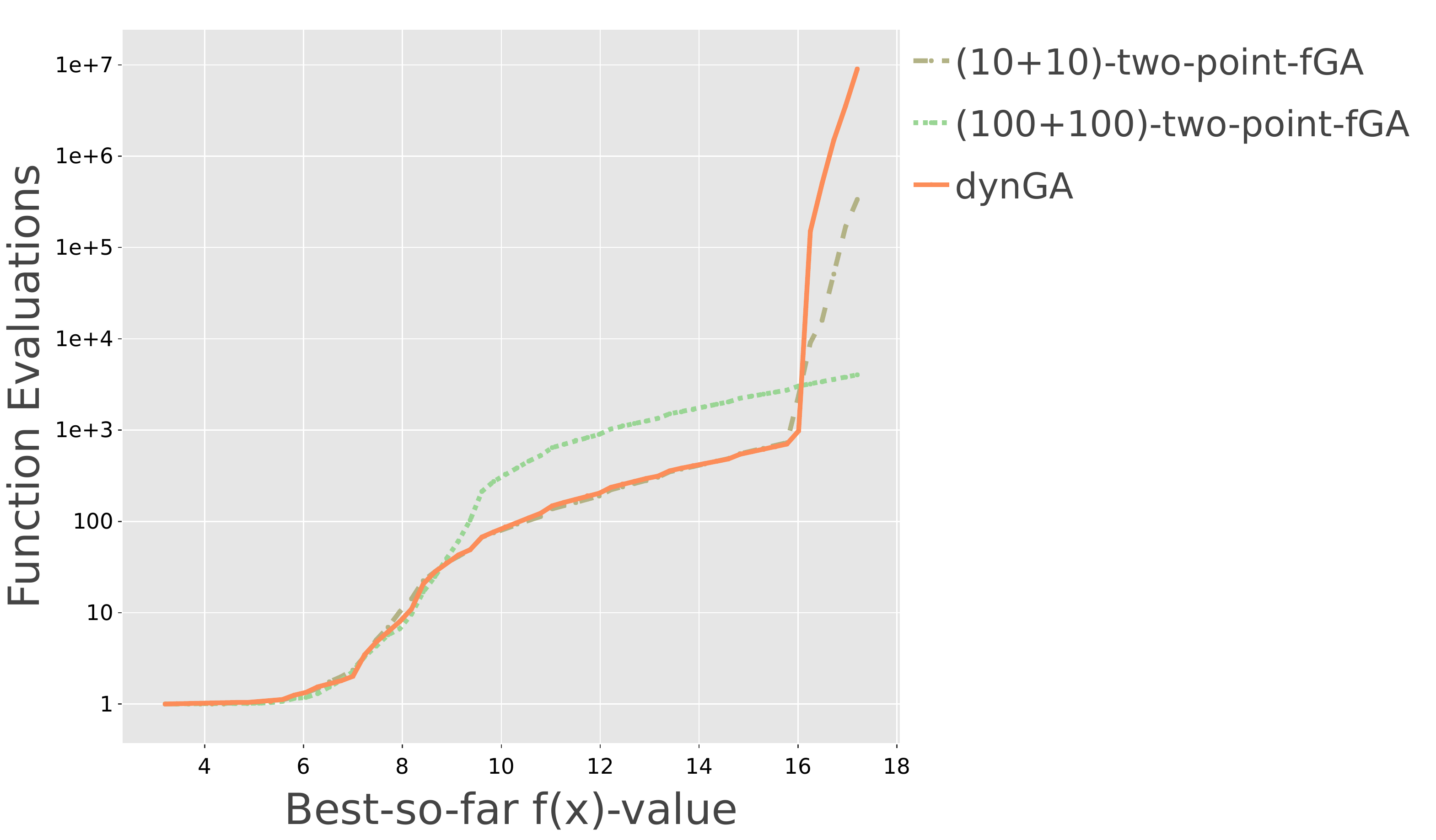}
 \vspace{-0.2cm}
\caption{Fixed-target ERTs of GAs on F24 in dimension $d=100$. The dynGA switches from the $(10+10)$-two-point-fGA to the $(100+100)$-two-point-fGA at the target $f(x)=15.81$. Results are from 100 independent runs. The figure is produced using the IOHprofiler tool~\cite{doerr2018iohprofiler}. }
\vspace{-0.2cm}
 \label{fig:F24}
\end{figure}

Figure~\ref{fig:F24} plots the fixed-target result of the best tested dynGA on F24, which uses a $(10+10)$-two-point-fGA at first and switches to a $(100+100)$-two-point-fGA afterward. By using a small population size $(10+10)$ initially, the dynGA indeed converges to the switch point fast, but it is trapped there and could not follow the original trend of the $(100+100)$ GA later. 

We do not solve this problem here, but it is interesting to spot this issue for future work. Concerning $A_1$, its performance at the switch point should be considered from different perspectives. If $A_1$ leads the dynAS policy into a local optimum, we should set the switch point earlier. Regarding $A_2$, it makes sense that $(100+100)$ can avoid being trapped for F24 with a large population size, but the question is how the algorithm handles local optima. If the algorithm obtains the ability to escape from local optima by means of the diversity of the population, we can expect to solve the dynAS by considering the initialization of $A_2$.
If the algorithm possesses powerful operators to escape from local optima, the method of formula~\ref{formula:pre-ERT} can still be useful to predict the performance of the dynAS policy. If the algorithm can avoid entering the local optima area but obtain the ability to escape from the area, we need to set the switch point before being trapped.

\subsubsection{A Successful  Case}
Although there are problems that the dynAS does not find better solvers as discussed, we gain improvement on most of the benchmark problems. 
Nevertheless, our goals are to obtain better results for problems and analyze the performance of GAs by applying the dynAS. In this section, we take the successful trial of F7 as an example to illustrate what we can achieve by using the informed dynAS.

\begin{figure}[htb]
 \includegraphics[width=0.45\linewidth, trim = 0mm 0mm 0mm 0mm, clip]{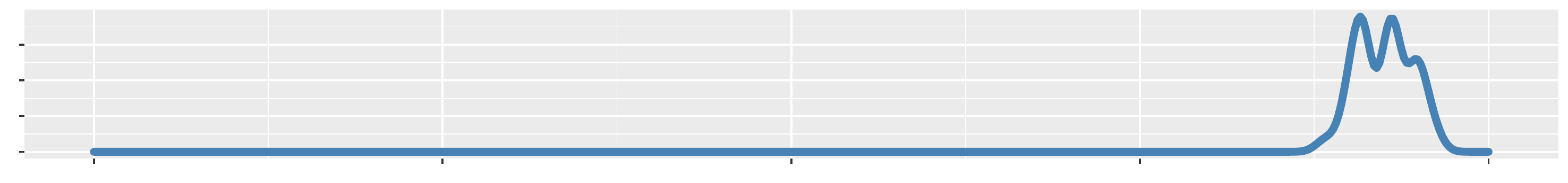}
 \includegraphics[width=0.6\linewidth, trim = 0mm 0mm 0mm 0mm, clip]{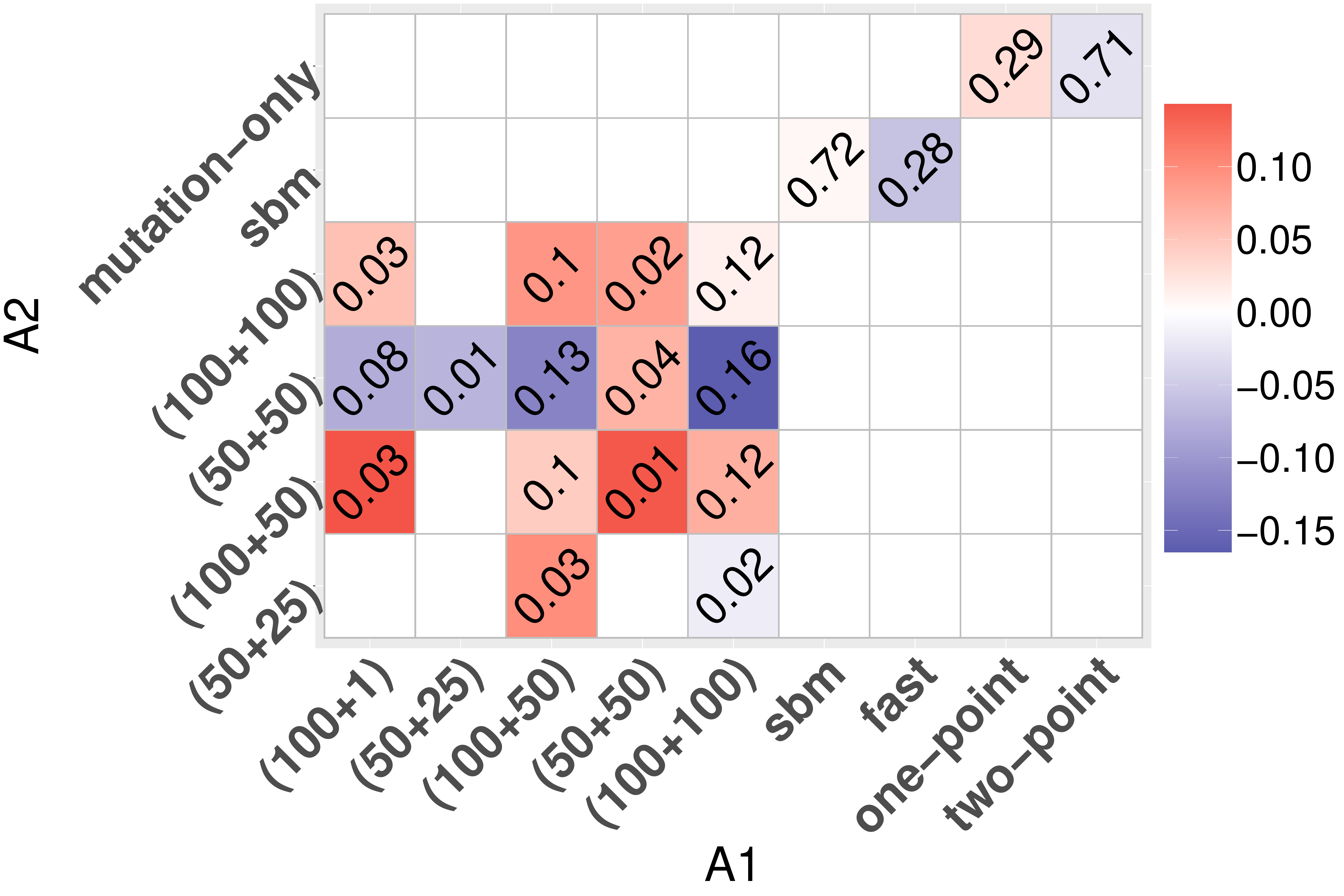}
  \vspace{-0.3cm}
 \caption{Averaged Relative ERTs of dynGAs with corresponding operator combination relative to the $sERT$ on F7 in dimension 100. X-axis and Y-axis indicate the operators selected by $A_1$ and $A_2$ respectively. Purple color indicates better solvers comparing to the BSA. Numbers on tiles show the frequency of the combination that appears among $100$ algorithms.\\
 The figure on the top of the tile plot is the distributions of $\phi_s$ of $100$ dynGAs, and values are scaled by $(\phi_s-\phi_m)/(\phi_f-\phi_m)$. $\phi_m$ is the minimal fitness of the problem.}
 \vspace{-0.4cm}
 \label{fig:F7-Operators}
\end{figure}

Figure~\ref{fig:F7-Operators} presents the frequencies of combinations of GAs and their corresponding relative ERTs comparing to the $sERT$ on F7. We observe that various GAs are selected for the dynAS policies, and the superior settings can be easily recognized.
According to Table~\ref{tab:TheoResult}, $(50+50)$~EA$_{>0}$ is the BSA for F7. Meanwhile, the dynGAs gain improvement in Figure~\ref{fig:F7-Operators} by using $(50+50)$~EA$_{>0}$ as $A_2$ . For $A_1$, $(100+100)$-two-point-GA is the one that can be useful for the dynGAs. Based on the observation, we expect that, for such a \onemax variant of epistasis, using two-point crossover can save function evaluations in the early stage, and a mutation-only GA will be the right choice for the later stage.

\begin{figure}[htb]
 \includegraphics[width=0.75\linewidth, trim = 0mm 0mm 0mm 0mm, clip]{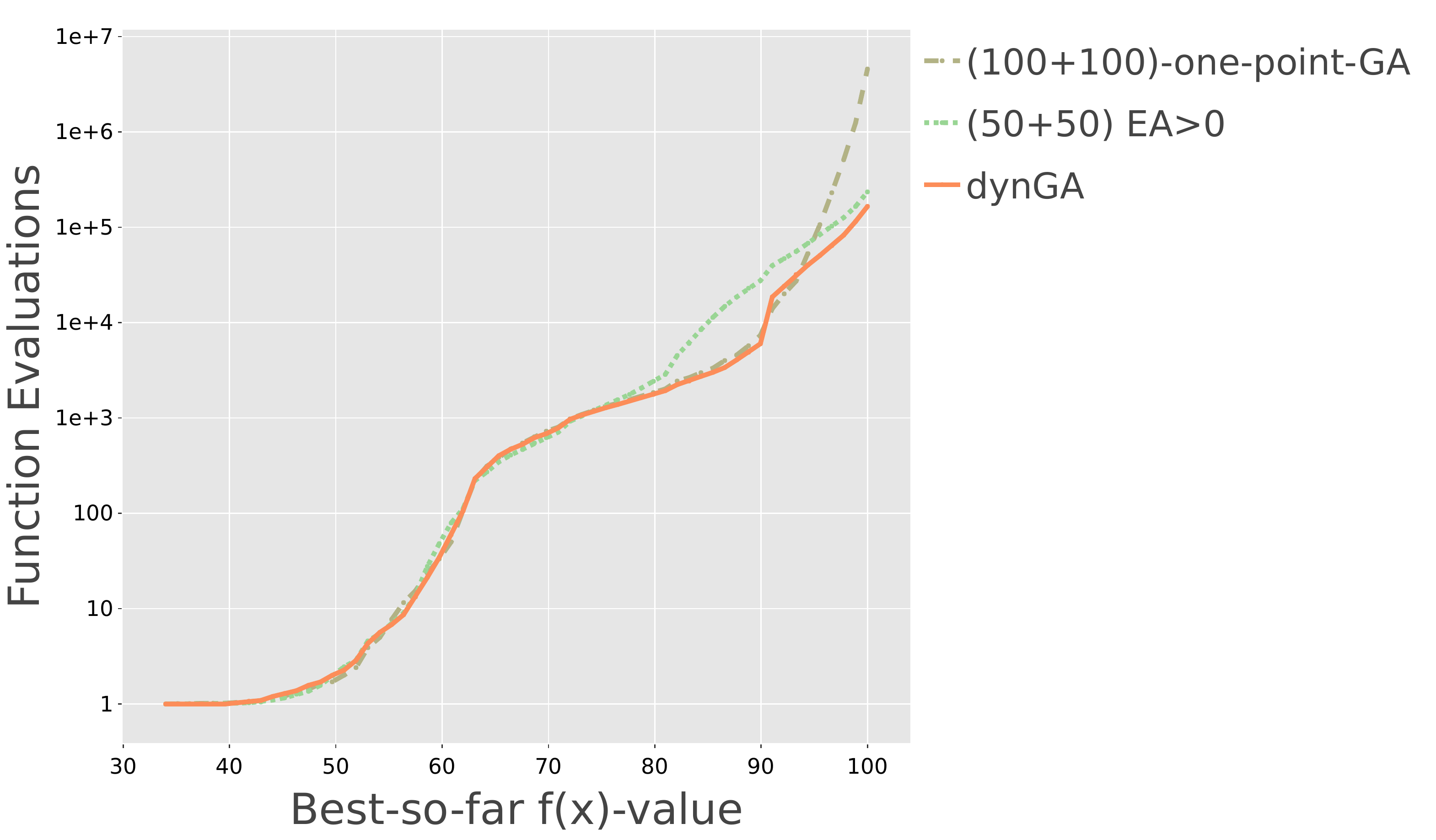}
 \vspace{-0.2cm}
 \caption{Fixed-target ERTs of GAs on F7 in dimension $d=100$. The dynGA switches from the $(100+100)$-one-point-GA to the $(50+50)$~EA$_{>0}$ at the target $f(x)=90$. Results are from 100 independent runs. The figure is produced using the IOHprofiler tool~\cite{doerr2018iohprofiler}. }
 \vspace{-0.5cm}
 \label{fig:F7}
\end{figure}

Additionally, we plot the fixed-target result of the best dynGA on F7 in Figure~\ref{fig:F7}. Interestingly, the $A_1$ of the best dynGA is using one-point crossover instead of two-point crossover. According to Figure~\ref{fig:F7-Operators}, we do not observe a significant improvement by using one-point crossover for $A_1$. By analyzing raw data, we find this advantage is hidden by averaging other dynAS policies. The distribution of $\phi_s$ (Figure~\ref{fig:F7-Operators}) shows two peaks around $90$ and $93$ respectively, but the performance of the dynAS policy deteriorates as $\phi_s > 90$ increases, though the theoretical prediction still indicates an improvement. This observation reflects the discussion in Sec~\ref{sec:localoptima} that the switching point should be chosen by considering the state of $A_1$.

\section{Conclusion and Future Work}\label{sec:conclusions}
We have investigated in this work possibilities to leverage existing benchmark data to derive switch-once dynamic algorithm selection policies. Our use-case was a family of genetic algorithms, applied to the 25 problems suggested in~\cite{doerr2020benchmarking,Yeppsn2020}. We first used the benchmark data to compute a hypothetical performance of the dynAS policies. We then executed the ones which showed the best improvement potential. Our experimental analysis confirmed the existence of combinations which outperform the best static algorithms. For the dynGAs that do not perform as expected, we could either explain the reasons or we offered a more fine-grained investigation of our dynAS approach.
We have also analyzed the role of the diversity of the candidate algorithms, the choice of the switch points, and of the local optima.

Moreover, we highlight the competitive GAs of stages of the optimization process for some problems. Applying uniform crossover can be helpful at the late stage of optimization for \leadingones, and the experimental result shows that we can gain improvement by switching to the $optimal$ crossover probability dynamically. Uniform crossover is useful at the late stage of optimization for \onemax, but the dynAS has not recognized this advantage for the \onemax variants with weighted variables, dummy variables, and neutrality. The dynGA gains improvement over the BSA of \oea for the \onemax variant of ruggedness by starting with the \oea and switching to the GA with uniform crossover. Oppositely, one-point and two-point crossover can accelerate the early optimization for the \onemax variant with epistasis, but the standard bit mutation with $p=1/n$ is a better choice for the late stage.

\textit{Understanding and Design of Algorithms.} The previous result on F3-6 has shown that we can not rely on dynAS to achieve better solvers when the potential of the set of algorithms is limited, but it can still help us understand how the different algorithms perform in the different stages of the optimization process. Such insights can facilitate the design of new algorithms on the one hand, and it can support theoretical analyses on the other.  

\textit{Performance Measures.} We have used in this work the ERT performance measure. 
Our results revealed that this cost measure has several drawbacks for the use within one-shot informed dynAS. Firstly, its value can be affected by the budget for the experiments with unsuccessful runs. For the second stage of the \textit{switch-ones} dynAS, if an algorithm cannot hit the target at the switching point at all runs, the later segment of formula~\ref{formula:pre-ERT} will not reflect its performance as $A_2$ accurately. Secondly, the ERT only reflects the performance with respect to the target. We can not utilize the performance before the algorithm hits the target by using it for the dynAS. We could mitigate these shortcomings by considering other measures such as the area under the empirical distribution function curve, which considers a set of targets and the fraction of successful runs.

\bibliographystyle{ACM-Reference-Format}
\bibliography{reference}

\appendix

\end{document}